\documentclass[acmsmall]{acmart}
\AtBeginDocument{%
  \providecommand\BibTeX{{%
    \normalfont B\kern-0.5em{\scshape i\kern-0.25em b}\kern-0.8em\TeX}}}

\setcopyright{acmcopyright}
\copyrightyear{2023}
\acmYear{2023}
\acmDOI{XXXXXXX.XXXXXXX}

\acmJournal{JACM}
\acmVolume{37}
\acmNumber{4}
\acmArticle{127}
\acmMonth{8}

\usepackage{pifont}
\newcommand{\cmark}{\ding{51}}%
\newcommand{\xmark}{\ding{55}}%
\usepackage{longtable}
\usepackage{tabularx}
\usepackage{paralist}
\usepackage{booktabs}

\def\aiqidel#1{\bgroup\markoverwith{\textcolor{blue}{\rule[0.5ex]{2pt}{1pt}}}\ULon{#1}}

\def\arkaitzdel#1{\bgroup\markoverwith{\textcolor{red}{\rule[0.5ex]{2pt}{1pt}}}\ULon{#1}}

\begin{document}

\title[A review of cross-lingual offensive language detection]{Cross-lingual Offensive Language Detection: A Systematic Review of Dataset, Approach and Challenge}

\author{Aiqi Jiang}
\email{a.jiang@qmul.ac.uk}
\affiliation{%
  \institution{Queen Mary University of London}
  \streetaddress{Mile End Road}
  \city{London}
  \country{UK}
  \postcode{E1 4NS}
}
\affiliation{%
  \institution{Heriot-Watt University}
  \city{Edinburgh}
  \country{UK}
  \postcode{EH14 4AS}
}

\author{Arkaitz Zubiaga}
\email{a.zubiaga@qmul.ac.uk}
\affiliation{%
  \institution{Queen Mary University of London}
  \streetaddress{Mile End Road}
  \city{London}
  \country{UK}
  \postcode{E1 4NS}
}

\renewcommand{\shortauthors}{A. Jiang and A. Zubiaga}

\begin{abstract}
The growing prevalence and rapid evolution of offensive language in social media amplify the complexities of detection, particularly highlighting the challenges in identifying such content across diverse languages.
This survey presents a systematic and comprehensive exploration of Cross-Lingual Transfer Learning (CLTL) techniques in offensive language detection in social media.
Our study stands as the first holistic overview to focus exclusively on the cross-lingual scenario in this domain. 
We analyse 67 relevant papers and categorise these studies across various dimensions, including the characteristics of multilingual datasets used, the cross-lingual resources employed, and the specific CLTL strategies implemented. 
According to ``what to transfer'', we also summarise three main CLTL transfer approaches: instance, feature, and parameter transfer. 
Additionally, we shed light on the current challenges and future research opportunities in this field. 
Furthermore, we have made our survey resources available online, including two comprehensive tables that provide accessible references to the multilingual datasets and CLTL methods used in the reviewed literature.
\end{abstract}




\maketitle



\textcolor{red}{\textbf{Disclaimer:} \textit{Due to the nature of this work, some examples may contain offensive text and hate speech.}}

\section{Introduction}


The prevalence of offensive language on social media platforms, such as Facebook and Twitter, has become a pressing concern in recent years. 
This surge can be attributed to the anonymity provided by these platforms and the lack of strict regulations and effective mechanisms to restrain such behaviour \cite{fortuna2018survey}.
While social platforms have fostered connections and bridged global distances, they have in turn enabled the spread of hate speech (HS) and various forms of offensive language \cite{chhabra2023literature}.
Offensive language broadly refers to expressions that may upset or annoy others,
while HS escalates beyond mere offensive behaviour to inciting discrimination, hostility, or violence against individuals or groups based on their race, religion, ethnicity, gender, or other identity factors \cite{rightforpeace}. 
HS is considered a human rights violation in many legal systems due to its potential to cause real-world harm and perpetuate discrimination against targeted groups \cite{unspahs2019}. 
Hence, there is a growing scholarly focus on developing approaches to mitigate HS in addition to offensive content.


The Natural Language Processing (NLP) community has proposed a variety of research techniques to detect offensive content based on both traditional machine learning and advanced neural network-based approaches.
While early work focused on monolingual settings (i.e. English), research in recent years has increasingly focused on tackling the problem in multilingual settings to broaden the ability to deal with more languages, a key requirement given the presence of diverse languages in social platforms \cite{pamungkas2021towards}.
This however generally hindered by the limited availability of labelled data and the high variability of offensive language across diverse cultures and languages \cite{plaza2023respectful}.

Cross-Lingual Transfer Learning (CLTL) emerges as a promising direction to mitigate challenges associated with data scarcity by leveraging domain knowledge from high- to low-resource languages \cite{pikuliak2021cross}.
Given its adaptability and compatibility with neural networks, CLTL has found successful applications across NLP tasks \cite{pan2010survey}. Some pioneering efforts have also integrated CLTL for offensive language detection in low-resource languages \cite{jiang2021cross,shi2022cross,zhou2023cross}. 
However, open challenges remain in building effective and generalisable cross-lingual models and understanding linguistic gaps.

Interest in cross-lingual offensive language detection has increased since the first work in 2018 \cite{mathur2018detecting}, but this is yet to be covered in a comprehensive survey. 
Existing surveys concentrate on monolingual detection or specific characteristics of offensive language, while the unique CLTL application in offensive language detection is seldom considered.
Relevant surveys exist \cite{pamungkas2021towards,chhabra2023literature}, but their scope goes beyond cross-lingual scenarios, also covering cross-domain and multimodal dimensions, and hence lacking our survey's focus on CLTL.


\subsection{Scope of the Survey}


Our objective is to provide a systematic review of existing literature on CLTL techniques for offensive language detection, and consequently, this survey discusses \textbf{67 papers} in this area.  
To make this scope explicit, we set out the following key objectives:

\begin{itemize}
    \item \textbf{Map the landscape} of multilingual datasets and cross-lingual resources used in offensive language detection, discussing aspects such as languages, label schemes, and availability.
    \item \textbf{Classify and analyse methods} by introducing a fine-grained taxonomy of CLTL approaches tailored to this task, distinguishing instance-, feature-, and parameter-level transfer.
    \item \textbf{Identify key challenges} that limit progress, with a focus on language diversity, dataset scarcity, cultural variation, and modelling limitations in low-resource scenarios.
    \item \textbf{Recommend future research directions} for developing more generalisable, robust, and ethically sound cross-lingual detection systems.
\end{itemize}

Furthermore, we make survey resources available online,\footnote{\url{https://github.com/aggiejiang/crosslingual-offensive-language-survey}} including two comprehensive tables that summarise multilingual datasets and CLTL methods used in surveyed papers.


\subsection{Survey Structure}

The rest of the paper is organised as follows. 
\S\ref{sec-background} describes the definitions of offensive language and the framework of cross-lingual detection, and distinguishes `cross-lingual' from related terminologies such as `multilingual'. Related surveys are also summarised. 
\S\ref{sec-survey-method} summarises our survey methodology. 
Then we summarise and analyse the multilingual datasets leveraged in surveyed papers in \S\ref{sec-dataset}. 
In \S\ref{sec-cl-resource}, we describe diverse linguistic resources and tools used in cross-lingual studies. 
\S\ref{sec-cl-method} discusses the three transfer levels of CLTL in offensive language detection, complemented by transfer strategies based on these layers.
\S\ref{sec-challenge} presents current challenges and future research directions in this area, concluding the survey in \S\ref{sec-conclusion}.

\section{Background and Related Work}
\label{sec-background}

\subsection{Cross-linguality in Offensive Language Detection}

\noindent\textbf{Task Definition and Workflow.}
The task focuses on identifying and categorising offensive or hateful text across languages. 
The objective of CLTL is to leverage the knowledge acquired from one language (source language with more resources) to enhance HS detection in another language (target language with less resources), especially when the labelled data in the target language is scarce or lacking \cite{pamungkas2021towards}.
The task workflow can be thought of as a four-stage process:

\begin{compactenum}
    \item \textbf{Data Preparation.}
    Gather text in both a source language with more labelled examples and a target language with fewer or no labels. The source data provides the foundation for learning, while the target data is where we ultimately want to detect offensive content.
    
    \item \textbf{Cross-lingual Training.}
    Use the source data alone, or combine it with any available target data, to train a model how to recognise offensive language patterns.

    \item \textbf{Model Adaptation.}
    If any labelled examples exist in the target language, fine-tune the model so it better reflects specific characteristics of the target language.

    \item \textbf{Detection on Target Language.}
    Apply the adapted model to unseen target language text to assign offensive/non-offensive labels.
\end{compactenum}

Formally, we represent the process as follows.  
Let $L_s$ and $L_t$ be the source and target languages, with datasets $D_s = \{X_s, Y_s\}$ and $D_t = \{X_t, Y_t\}$, where $X$ denotes text and $Y \in \{0,1\}$ denotes non-offensive (0) or offensive (1) labels. 
The cross-lingual task seeks a function $f: F \rightarrow Y_t$ mapping a feature space $F$ to the target label space.
In the training stage, it can use only $D_s$ or both $D_s$ and $D_t$:

\begin{itemize}
    \item \textbf{Source-only training:} Train $M_s$ on $D_s$ to learn $f_s: F_s \rightarrow Y$, then adapt $f_s$ to $f_t: F_t \rightarrow Y$.
    \item \textbf{Joint training:} Train $M_{st}$ on $D_s \cup D_t$ to learn $f_{st}: F_{st} \rightarrow Y$ based on joint feature space $F_{st}$.
\end{itemize}

\noindent If $Y_t$ is available, fine-tune $f_t$ (or $f_{st}$) on full or few-shot $D_t$, producing $M_t$ for prediction:
\[
\hat{y}_t = f_t(D_t) \quad \text{or} \quad \hat{y}_t = f_{st}(D_t).
\]

\noindent\textbf{Cross-lingual \& Multilingual \& Code-mixing.}
While ``cross-lingual learning'' typically denotes the application of CLTL techniques, the term ``multilingual learning'' is sometimes used interchangeably \cite{pikuliak2021cross}.
Some multilingual data may contain code-mixed content, i.e. a mixture of languages in the same sentence \cite{ibrohim2019translated}.
Although some researchers draw parallels between these terms and the relationship between transfer learning and multi-task learning \cite{wang2015transfer}, we adopt a broader perspective on cross-lingual learning.
We define cross-lingual learning as the process of transferring knowledge between different languages, a concept that inherently subsumes multilingual learning.

\vspace{-1mm}
\subsection{Related Surveys}
\label{sec-related-survey}

We found 18 surveys covering offensive language detection from different angles.

\vspace{0.8mm}
\noindent\textbf{General Offensive Language Detection.}
Surveys on offensive language detection provide general insights into the field \cite{schmidt2017survey,fortuna2018survey,jahan2023systematic,mishra2019tackling}. 
Key works by \citet{schmidt2017survey} and \citet{fortuna2018survey} stand out as the two most cited reviews in the NLP community, providing definitions of HS and related terms, summarising previous methods, and discussing the existing limitations and challenges.
Some of these surveys focus on in-depth discussions of specific aspects when reviewing studies in HS detection, such as data sources \cite{alkomah2022literature,vidgen2020directions,poletto2021resources}, methodologies \cite{alkomah2022literature,ayo2020machine,demilie2022detection}, and challenges \cite{ayo2020machine,kiritchenko2021confronting,yin2021towards,garg2023handling}.
Additionally, several surveys have also delved into related areas like cyberbullying \cite{yi2023session} and the intersection of HS and fake news \cite{demilie2022detection}, and others review ethical considerations in the development of HS detection \cite{vidgen2020directions,kiritchenko2021confronting}.


\vspace{0.8mm}
\noindent\textbf{Cross-lingual and Multilingual Offensive Language Detection.}
Several surveys focus on studies in single but non-English languages, such as Arabic \cite{husain2021survey,al2019detection}, Indian \cite{dowlagar2021survey} and Ethiopian \cite{demilie2022detection} languages.
Surveys for HS detection also provide a brief summary of multilingual or cross-lingual contributions \cite{poletto2021resources,al2019detection,yin2021towards}.
However, systematic reviews focusing explicitly on cross-lingual perspectives remain scarce \cite{pamungkas2021towards,chhabra2023literature}.
\citet{pamungkas2021towards} provide an overview of datasets and approaches in cross-domain and cross-lingual settings. 
\citet{chhabra2023literature} conduct a comprehensive review of multimodal and multilingual HS detection, including benchmark datasets, state-of-the-art methods, challenges, and evaluation techniques.


\vspace{0.8mm}
\noindent\textbf{Strength of our survey.} 
In contrast, our work presents:

\begin{itemize}
    \item \emph{Fine-grained taxonomy of CLTL approaches} for offensive language detection, structured into instance-, feature-, and parameter-level transfer, each with detailed strategy categorisation;
    \item \emph{Comprehensive dataset library} including rich metadata (e.g., language families, annotation schemes, abuse types, availability, etc.);
    \item \emph{In-depth analysis of challenges} unique to cross-lingual offensive language detection from aspects of languages, data sources and approaches;
    \item \emph{Future-oriented perspective} that synthesises promising research directions and provides recommendations for future research.
\end{itemize}

\section{Survey Methodology}
\label{sec-survey-method}

\begin{figure}[!ht]
  \centering
  \includegraphics[width=0.99\linewidth]{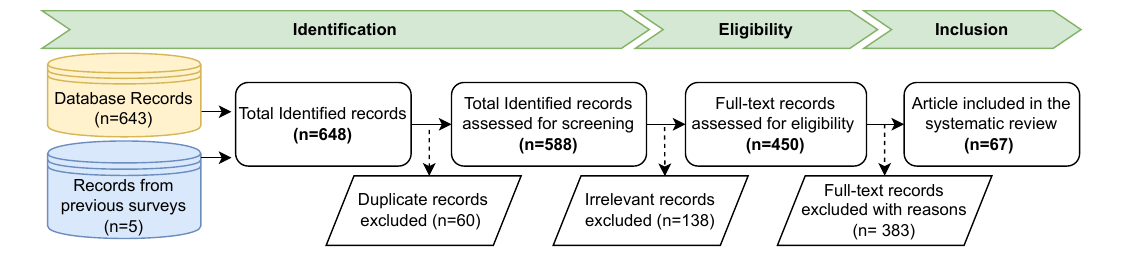}
  \caption{PRISMA flowchart showing the phases of the selection of research articles in this review.}
  \label{fig:prisma}
\end{figure}

We adopt the structured PRISMA framework \cite{page2021prisma} for our systematic review to ensure both replicability and transparency. A detailed flowchart of our search protocol is provided in Figure \ref{fig:prisma}, which illustrates the progression of publications through different phases of our selection process.

\vspace{0.8mm}
\noindent\textbf{Search Databases.}
We retrieve publications from scientific databases including
ACL Anthology,\footnote{\url{https://aclanthology.org/}} 
Google Scholar,\footnote{\url{https://scholar.google.com/}} 
and DBLP Computer Science Bibliography.\footnote{\url{https://dblp.org/}}

\vspace{0.8mm}
\noindent\textbf{Keyword Selection.}
We create a list of 99 pairwise combinations of words in (i) and (ii):
(i) 9 keywords related to offensive language, such as ``hate'', ``abus*'', ``offens*'', ``toxic'', ``cyberbully'', ``gender'', ``sexis*'', ``misogyn*'', ``rac*'';
(ii) 11 keywords related to multilingual and cross-lingual scenarios, such as ``cross-lingual'', ``multilingual'', ``low resource'', ``under resource'', ``cross language'', ``between language'', ``cross linguistic'', ``transfer'', ``code switch'', ``zero shot'', ``few shot''.

\vspace{0.8mm}
\noindent\textbf{Search Process.}
In addition to studies matching the 99 keyword pairs above, 
we also identify and include 5 additional studies cited in relevant surveys.
This search was conducted on July 29th, 2023.

\vspace{0.8mm}
\noindent\textbf{Eligibility Criteria.}
Following removal of duplicates and irrelevant papers, the remainder 455 publications were checked against the inclusion / exclusion criteria in Table \ref{tab:eligibility}.



\begin{table}[ht!]
    \centering
    \fontsize{6.7}{7.5}\selectfont
    \begin{tabular}{p{6.7cm}|p{6.6cm}}
        \toprule
        \textbf{Include} & \textbf{Exclude} \\
        \midrule
        (i) Describe cross-lingual approaches to detect general abusive content (toxic, hate, etc.)  & (i) Study transfer across dimensions other than languages (e.g. platforms, domains, datasets) \\
        (ii) Describe cross-lingual approaches to detect subtypes of abusive content (e.g. sexism, racism)  & (ii) Study low-resource languages without knowledge transfer approaches \\
        (iii) Describe multilingual approach with knowledge transfer between languages &  (iii) Relevant studies from competition reports and papers \\ 
        (iv) Study text classification cases in cross-lingual HS scenarios &  (iv) Master or PhD dissertations \\
        \bottomrule
    \end{tabular}
    \normalsize 
    \caption{Eligibility criteria for inclusion / exclusion studies.}
    \label{tab:eligibility}
\end{table}

\begin{figure}[ht!]
  \centering
  \includegraphics[width=0.6\linewidth]{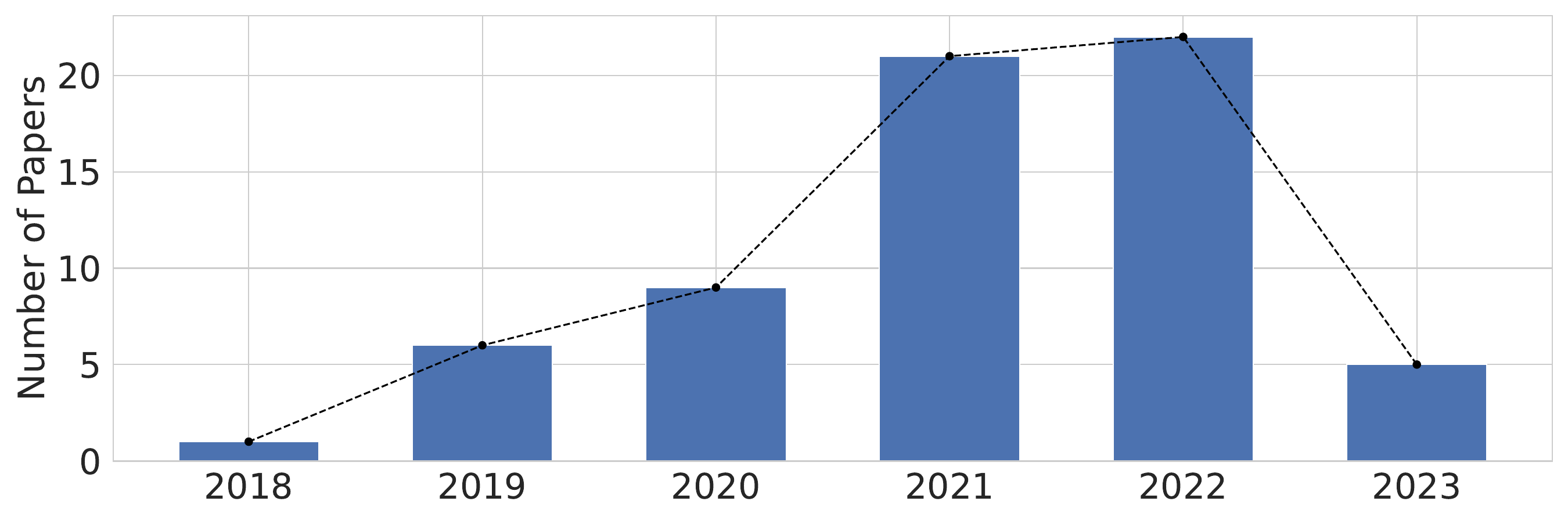}
  \caption{Publications per year up to July 2023.}
  \label{fig:paper}
\end{figure}

\vspace{0.8mm}
\noindent\textbf{Summary.}
In the end we obtained 67 relevant papers, with their temporal distribution shown in Figure \ref{fig:paper} and showing a growing trend from the first publication in 2018 to 2022.

\section{Multilingual Hate Speech Datasets}
\label{sec-dataset}


We investigate 82 datasets utilised in existing surveyed studies on cross-lingual HS detection across different aspects. We summarise these datasets and their characteristics in Table \ref{tab:datasets}.

\begin{center}
    \fontsize{6.7}{7.5}\selectfont
    \begin{longtable}{p{0.3cm}cp{2.6cm}p{2.6cm}p{1.5cm}p{1cm}p{0.4cm}p{0.4cm}ccc}
    \toprule
        \textbf{Ref} & \textbf{Year} & \textbf{Topic} & \textbf{Source} & \textbf{Language} & \textbf{Size} & \textbf{Label}   & \textbf{\#Cit} & \textbf{CM?} & \textbf{Avail?}  \\
        \midrule
        \endfirsthead
        \multicolumn{10}{l}{\textbf{[Continued table]}} \\
        \midrule
        \endhead

        \midrule
        \endfoot

        \bottomrule
        \\
        \caption{Summary of datasets used in cross-lingual offensive language detection.
        ``\#Cit'' = ``number of citations'', ``CM?'' = ``whether or not the dataset is code-mixed'', ``Avail?'' = ``dataset availability'', and ``Label'' denotes: (1) binary labels, (2) fine-grained offense category, (3) attack target, and (4) intensity score.} 
        \label{tab:datasets}
        \endlastfoot
        \cite{paul2023covid} & 2023 & cyberbullying & Twitter & en, hi & 22k &  1 & <50 &\cmark & \xmark\\
        \cite{jeong2022kold} &2022& offense & NAVER-news, Youtube & ko & 40,429 &  1,2,3 & <10 & \xmark  &\cmark \\
        \cite{jiang2022swsr} &2022& Sexism & Weibo &  zh & 8,969 &  1,2,3 & <50 & \xmark  &\cmark \\
        \cite{salaam2022offensive} &2022& offense & - & en, fr, de, es  & 37,221 &  1 & <10 & \cmark  &\cmark \\
        \cite{deng2022cold} & 2022 & offense & Weibo, Zhihu & zh & 37,480 &  1,2 & <50 &\xmark & \cmark\\
        \cite{gupta2022multilingual} & 2022 & abuse & ShareChat & hi, kn, ml, ta, te  & 92,881  &  1 & <10 &\cmark & \cmark\\   
        \cite{chakravarthi2022dravidiancodemix} & 2022 & offense & Youtube &  ta  & 60k  &  1,3 & <50 &\cmark & \cmark\\    
        \cite{arango2022resources} & 2022 & offense & Twitter &  es  & 9,834  &  2 & <10 &\xmark & \cmark\\
        \cite{bhattacharya2020developing} & 2022 & aggressiveness, misogyny & Youtube &  bn  & 25k  &  1 & <100 &\xmark & \cmark\\    
       \cite{kumar2022multi} & 2022 & cyberbullying & Facebook, Twitter &  en, hi  & 6,500  &  1 & <50 &\cmark & \xmark\\ 
       \cite{pavlopoulos-etal-2021-semeval} & 2021 & toxicity & Civil Comments &  en  & 10,629  &  1,2 & <100 &\xmark & \cmark\\  
       \cite{burtenshaw2021dutch} & 2021 & toxicity & Ask.fm &  nl  & 10,189  &  2 & <10 &\xmark & \xmark\\       
       \cite{hande2021offensive} & 2021 & offense & Youtube &  kn, ml, ta  & 71,691  &  1,3 & <20 &\cmark & \cmark\\
       \cite{mathew2021hatexplain} & 2021 & hate speech & Gab, Twitter &  en  & 20,148  &  1,2 & <500 &\xmark & \cmark\\ 
       \cite{Romim2020HateSD} & 2021 & hate speech & Facebook, Youtube &  bn  & 30k  &  1 & <50 &\xmark & \xmark\\
       \cite{modha2021hasoc} & 2021 & hate speech, offense & Twitter &  en, hi, mr  & 13,755  &  1,4 & <100 &\xmark & \cmark\\
       \cite{amjad2022urdu} & 2021 & abuse & Twitter &  ur  & 8,400  &  1 & <50 &\xmark & \cmark\\
       \cite{vidgen2020learning} & 2021 & hate speech & open-source platform &  en  & 41,255  &  1,2 & <100 &\xmark & \cmark\\
       \cite{gaikwad2021cross} & 2021 & offense & Twitter &  mr  & 2499  &  1,2,3 & <50 &\xmark & \cmark\\
       \cite{khan2021hate} & 2021 & hate speech, offense & Twitter &   roman ur  & 5k  &  1 & <50 &\xmark & \xmark\\
       \cite{mubarak-etal-2021-arabic} & 2021 & offense& Twitter &  ar  & 10k  &  1,2 & <250 &\xmark & \cmark\\
       \cite{pitenis2020offensive} & 2020 & offense & Twitter &  el  & 4,779  &  1 & <250 &\xmark & \cmark\\
       \cite{charitidis2020towards} & 2020 & hate speech& Twitter & en, fr, de, el, es  & 264,035  &  1 & <50 &\xmark & \cmark\\
       \cite{glavas2020xhate} & 2020 & hate speech, abuse & Facebook, Fox news, Twitter, Wikipedia & sq, hr, en, de, ru, tr  & 109,955  &  1 & <50 &\xmark & \cmark\\
       \cite{Shekhar2020AutomatingNC} & 2020 & offense & 24sata, Eesti Ekspress, Veˇcernji List & hr, et  & 62.6m  &  - & <10 &\xmark & \xmark\\
       \cite{akhter2020automatic} & 2020 & offense, abuse& Youtube & Roman ur, ur  & 12,171  &  1 & <100 &\xmark & \cmark\\
       \cite{rizwan2020hate} & 2020 & hate speech, offense & Twitter & Roman ur  & 10,012  &  1,2 & <50 &\xmark & \cmark\\       
       \cite{mandl2021hasoc} & 2020 & offense & Twitter, Youtube & en, de, hi, ml, ta  & 15,047  &  1,2 & <250 &\cmark & \cmark\\       
       \cite{sanguinetti2020haspeede} & 2020 & hate speech & Twitter & it  & 12,081  &  1 & <100 &\xmark & \cmark\\ 
       \cite{kennedy2020contextualizing} & 2020 & hate speech & Gab & en  & 27,655  &  1 & <250 &\xmark & \cmark\\ 
       \cite{ccoltekin2020corpus} & 2020 & offense & Twitter & tr  & 36,232  &  1,3 & <250 &\xmark & \cmark\\
       \cite{kumar2020evaluating} & 2020 & aggressiveness & Youtube & bn, en, hi  & 15k  &  1 & <250 &\xmark & \cmark\\
       \cite{sigurbergsson-derczynski-2020-offensive} & 2020 & abuse, offense & Facebook, Reddit, Twitter & da  & 3,600  &  1,3 & <250 &\xmark & \cmark\\
       \cite{hande-etal-2020-kancmd} & 2020 & offense & Youtube & kn  & 7,671  &  1,3 & <100 &\cmark & \xmark\\
       \cite{zampieri2020semeval} & 2020 & offense & Twitter & ar, da, en, el, tr  & 9m  &  1,3 & <500 &\xmark & \cmark\\               
       \cite{ljubevsic2019frenk} & 2019 & offense & Facebook & en, sl  & 22,877  &  1,2 & <50 &\cmark & \xmark\\
       \cite{arata2019study} & 2019 & cyberbullying & Twitter & ja  & 4,096  &  1 & <10 &\xmark & \xmark\\
       \cite{arango2019hate} & 2019 & hate speech & Twitter & en  & 14,949  &  1 & <250 &\xmark & \cmark\\
       \cite{ibrohim-budi-2019-multi} & 2019 & abuse, hate speech & Twitter & id  & 5,561  &  1,2,3 & <250 &\xmark & \cmark\\
       \cite{mulki2019hsab} & 2019 & abuse, hate speech, toxicity & Twitter & at  & 5,846  &  1 & <250 &\xmark & \xmark\\  
       \cite{s19214654} & 2019 & hate speech & Twitter & es  & 6k  &  1 & <100 &\xmark & \cmark\\ 
       \cite{fortuna-etal-2019-hierarchically} & 2019 & hate speech, offense & Twitter & pt  & 5,668  &  1 & <100 &\xmark & \cmark\\
       \cite{poletto2021resources} & 2019 & cyberbullying, toxicity & Twitter & pl  & 11,041  &  1,2 & <50 &\xmark & \cmark\\
       \cite{strus2019germeval} & 2019 & offense & Twitter & de  & 7,025  &  1,2 & <250 &\xmark & \cmark\\
       \cite{basile2019semeval} & 2019 & aggressiveness& Twitter & en  & 19,600  &  1,3 & <750 &\xmark & \cmark\\
       \cite{zampieri2019semeval} & 2019 & offense & Twitter & en, es & 14k  &  1,3 & <700 &\xmark & \cmark\\
       \cite{mandl2019overview} & 2019 & hate speech, offense & Twitter, Facebook & en, de, hi & 17,657  &  1,2,3 & <100 &\xmark & \cmark\\
       \cite{chung2019conan} & 2019 & hate speech & Twitter & en, fr, it & 4,078  &  1,2 & <250 &\xmark & \cmark\\
       \cite{ousidhoum2019multilingual} & 2019 & hate speech & Twitter & ar, en, fr & 13,014  &  1,2,3 & <250 &\xmark & \cmark\\
       \cite{chiril2019multilingual} & 2019 & sexism, misogyny & Twitter & fr & 3,085  &  1 & <50 &\xmark & \xmark\\
       \cite{bosco2018evalita} & 2018 & hate speech & Facebook, Twitter & en, it & 4k  &  1 & <250 &\xmark & \cmark\\       
       \cite{gibert2018hate} & 2018 & hate speech & Stormfront & en & 10,568  &  1 & <500 &\xmark & \cmark\\
       \cite{ptaszynski2018cyberbullying} & 2018 & cyberbullying & Formspring.me & en & 12,772 &  1 & <50 &\xmark & \xmark\\
       \cite{alakrot2018dataset} & 2018 & abuse, offense & Youtube & ar & 167,549 &  1 & <100 &\xmark & \cmark\\      
       \cite{bohra-etal-2018-dataset} & 2018 & hate speech & Twitter & en, hi & 4,575 &  1 & <250 &\cmark & \cmark\\ 
       \cite{elsherief2018peer} & 2018 & hate speech & Twitter & en & 27,330 &  1,2  & <250 &\xmark & \cmark\\ 
       \cite{fersini2018evalita} & 2018 & misogyny& Twitter & en, it & 20k &  1,2,3  & <250 &\xmark & \cmark\\        
       \cite{mathur2018offend} & 2018 & abuse, hate speech, offense& Twitter & en, hi & 17,698 &  1,2  & <100 &\xmark & \cmark\\ 
      \cite{kumar2018aggression} & 2018 & aggressiveness & Facebook, Twitter & en, hi & 39k &  1,2,3  & <250 &\cmark & \cmark\\ 
      \cite{mathur2018detecting} & 2018 & offense& Twitter & en, hi & 3,679 &  1,2  & <250 &\cmark & \xmark\\ 
      \cite{sanguinett2018hate} & 2018 & aggressiveness, hate speech, offense & Twitter & it & 6k & 2,4 & <250 &\xmark & \cmark\\ 
      \cite{founta2018large} & 2018 & abuse, hate speech & Twitter & en & 80k & 1,2 & <550 &\xmark & \cmark\\
      \cite{wiegand2018overview} & 2018 & hate speech, offense & Twitter & de & 8,541 & 1,2 & <500 &\xmark & \cmark\\
      \cite{fersini2018ibereval} & 2018 & misogyny & Twitter & en, es & 8,115 & 1,2,3 & <250 &\xmark & \cmark\\
      \cite{golbeck2017harassment} & 2017 & harassment, offense, racism & BlockTogether, Twitter & en & 35k & 1 & <100 &\xmark & \cmark\\
      \cite{8004907} & 2017 & offense & Twitter & ru & 493 & 1,2 & <50 &\xmark & \xmark\\
      \cite{gao2017detecting} & 2017 & hate speech & Fox news & en & 1,528 & 1 & <250 &\xmark & \cmark\\
      \cite{wulczyn2017ex} & 2017 & hate speech & Wikipedia & en & 115,737 & 1 & <750 &\xmark & \cmark\\      
      \cite{rogers2017offensive} & 2017 & hate speech, offense & g1.globo.com & pt & 1,250 & 1 & <100 &\xmark & \cmark\\     
      \cite{ross2017measuring} & 2017 & hate speech, offense & Twitter & de & 541 & 1,2,4 & <500 &\xmark & \cmark\\      
      \cite{alfina2017hatedataset} & 2017 & hate speech & Twitter & id & 713 & 1 & <250 &\xmark & \cmark\\      
      \cite{Fortuna2017AutomaticDO} & 2017 & hate speech & Twitter & pt & 5,668 & 1,2 & <50 &\xmark & \cmark\\
      \cite{bretschneider2017detecting} & 2017 & hate speech, offense & Facebook & de & 5,836 & 1,3 & <100 &\xmark & \cmark\\
      \cite{davidson2017hatedataset} & 2017 & offense & Twitter & en & 24,802 & 1,2 & <2200 &\xmark & \cmark\\
      \cite{mubarak2017abusive} & 2017 & abuse, hate speech, offense & Twitter & ar & 32k & 1 & <500 &\xmark & \cmark\\
      \cite{waseem2016hate} & 2016 & hate speech & Twitter & en & 16,914 & 1 & <1550 &\xmark & \cmark\\
      \cite{van2015automatic} & 2015 & cyberbullying & Ask.fm & nl & 85,462 & 2,4 & <250 &\xmark & \xmark\\
     \cite{dybala2010machine} & 2010 & cyberbullying & informal websites of Japanese secondary schools & ja & 2,999 & 1 & <100 &\xmark & \xmark\\
    \end{longtable}
\end{center}

\noindent\textbf{Topics.}
The topic distribution among datasets shows that ``offensiveness'' and ``hate speech'' are the two most frequently addressed topics, accounting for 34.23\% and 32.43\% respectively. 
Some cover general offense, such as abusiveness, toxicity, cyberbullying, harassment and aggressiveness. 
In addition, some datasets narrow their scope to targeted issues, like sexism, misogyny and racism.


\vspace{0.8mm}
\noindent\textbf{Data Sources.}
Most studies collect data from social media such as Twitter (now X) and Facebook. 
Twitter encompasses 47.19\% of the datasets, followed by Youtube (11.24\%) and Facebook (10.11\%).
While these mainstream platforms dominate, news websites like Fox News and NAVER news, and open forums like Reddit and 2ch, also contribute to multilingual datasets.
Sources like Weibo (Chinese), 2ch (Russian), g1.globo (Brazilian Portuguese) and Eesti Ekspress (Estonian) represent non-English speaking regions as more culturally specific platforms.
Some datasets originate from platforms known for controversial content or specific user bases, such as Stormfront and alt-right websites. 


\begin{figure}[h]
  \centering
  \includegraphics[width=1.01\linewidth]{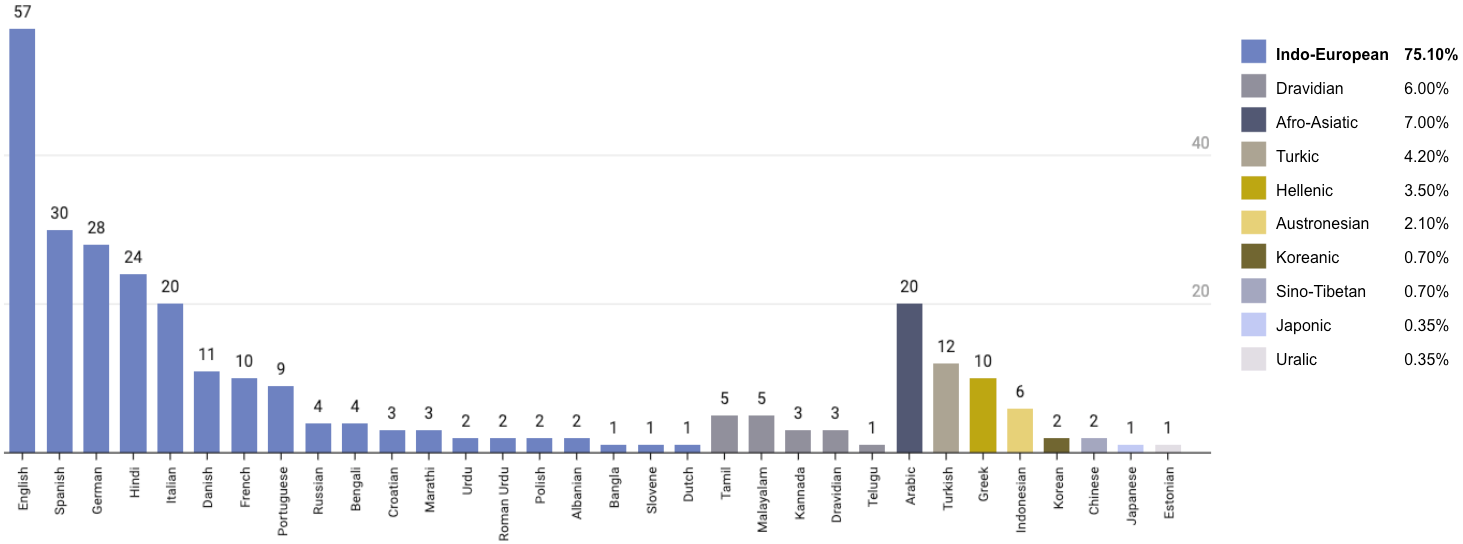}
  \caption{Distribution of languages and language families covered in the datasets.}
  \label{fig:lang-family}
\end{figure}

\vspace{0.8mm}
\noindent\textbf{Languages and Families.}
Datasets cover 32 languages across 10 language families, where one dataset may contain more than one language or family. We show the distribution of languages and families in Figure \ref{fig:lang-family}.
It highlights a representative language of English, accounting for 25.95\% of the data, followed by Hindi at 9.16\%, and German at 6.87\%. 
Notably, it exhibits a relatively balanced representation across several languages, such as Spanish, Arabic, Italian, French, Portuguese and Turkish, each ranging from around 2\% to 5\%. 
It also reveals a strong emphasis on Indo-European languages in HS studies, complemented by a notable presence of Afro-Asiatic languages (represented primarily by Arabic), with other language families being less represented.
Some datasets (14.6\%) are also dedicated to code-mixed content.

\vspace{0.8mm}
\noindent\textbf{Dataset Size.}
Over half of the datasets, accounting for 51.20\%, contain $10^4$-$10^5$ instances, closely followed by 35.40\% that have instances between $10^3$ and $10^4$. 
Besides, large datasets are rare, with only one exceeding $10^7$, and two between $10^6$ and $10^7$.
The majority of datasets fall within the smaller size ranges, which indicates the challenges in collecting and manually annotating large-scale datasets, as well as model generalisability for cross-lingual HS.

\vspace{0.8mm}
\noindent\textbf{Data Labelling and Distributions.}
According to diverse annotation schemes, four different types of labels are often employed: (i) binary labels, (ii) fine-grained categories of offensive content, (iii) attack targets, and (iv) intensity scores. 
A majority of datasets, 76 in total, utilise the simpler binary labelling, of which 36 solely provide binary labels.
There are also datasets combining binary labels with either fine-grained categories (20), attack targets (9), or intensity scores (1).
A smaller number of datasets (10) combine binary labels, fine-grained categories, and attack targets. 
Interestingly, only four datasets consider intensity scores, suggesting that quantifying the severity of offensive content is less common in current research. 


\vspace{0.8mm}
\noindent\textbf{Dataset Availability.}
We find 66 out of 82 datasets (80\%) are publicly accessible, with the other 16 unavailable, in some cases with the possibility of contacting the authors to request the data. 


\begin{table*}[ht!]
    \centering
    \fontsize{6.5}{7.5}\selectfont
    \begin{tabular}{p{2.5cm}p{5cm}cp{1.8cm}p{1.3cm}c}
        \toprule
        \textbf{Name} & \textbf{Description} & \textbf{Year} & \textbf{Topic} & \textbf{Language} & \textbf{\#Teams}\\
        \midrule 
        
        \href{https://projects.cai.fbi.h-da.de/iggsa/}{GermEval} \cite{wiegand2018overview} & Identification of Offensive Language & 2018 & offense & de & 20 \\ \addlinespace[1.5pt]
        \href{https://sites.google.com/view/ibereval-2018}{AMI@IberEval} \cite{fersini2018ibereval} & Automatic Misogyny Identification & 2018 & misogyny & en, es & 11 \\\addlinespace[1.5pt]
        \href{https://amievalita2018.wordpress.com}{AMI@Evalita} \cite{fersini2018evalita} & Automatic Misogyny Identification & 2018 & misogyny & en, it & 16 \\\addlinespace[1.5pt]
        \href{http://www.di.unito.it/~tutreeb/haspeede-evalita18/#}{HaSpeeDe@Evalita} \cite{bosco2018evalita} & Hate Speech Detection & 2018 & hate speech & it & 9 \\\addlinespace[1.5pt]
        \href{https://sites.google.com/view/trac1/shared-task?authuser=0}{TRAC-1} \cite{kumar2018aggression} & Aggression Identification of Hindi-English Code-mixed Data & 2018 & aggressiveness & en, hi & 30\\\addlinespace[1.5pt]
        \href{https://competitions.codalab.org/competitions/19935}{HatEval@SemEval} \cite{basile2019semeval} & Multilingual Detection of Hate Speech Against Immigrants and Women in Twitter & 2019 & hate speech & en, es & 74 \\\addlinespace[1.5pt]
        \href{https://hasocfire.github.io/hasoc/2019/index.html}{HASOC@FIRE} \cite{mandl2019overview} & Offensive Content Identification in Indo-European Languages & 2019 & hate speech, offense & en, hi, de & 37 \\\addlinespace[1.5pt]
        \href{https://fz.h-da.de/iggsa/}{Task2@GermEval} \cite{strus2019germeval} & Identification of Offensive Language & 2019 & offense & de & 13 \\\addlinespace[1.5pt]
        \href{https://github.com/ptaszynski/cyberbullying-Polish}{Task6@PolEval} \cite{ptaszynski2019poleval} & Automatic Cyberbullying Detection in Polish Twitter & 2019 & cyberbullying & pl & 9 \\\addlinespace[1.5pt]
        \href{https://sites.google.com/site/offensevalsharedtask/offenseval-2020}{OffensEval@SemEval} \cite{zampieri2020semeval} & Multilingual Offensive Language Identification in Social Media & 2020 & offense & en, ar, da, el, tr & 145 \\\addlinespace[1.5pt]
        \href{https://hasocfire.github.io/hasoc/2020/}{HASOC@FIRE} \cite{mandl2021hasoc} & Offensive Language Identification in Tamil, Malayalam, Hindi, English and German & 2020 & hate speech, offense  & en, hi, de, ta, ml & 40+ \\\addlinespace[1.5pt]
        \href{http://www.di.unito.it/~tutreeb/haspeede-evalita20/index.html}{HaSpeeDe@Evalita} \cite{sanguinetti2020haspeede} & Hate Speech Detection & 2020 & hate speech & it & 14 \\\addlinespace[1.5pt]
        \href{https://sites.google.com/view/trac2/home}{TRAC-2} \cite{kumar2020evaluating} & Aggression and Gendered Aggression Identification & 2020 & aggressiveness & en, hi, bn & 19 \\\addlinespace[1.5pt]
        \href{https://www.kaggle.com/c/jigsaw-multilingual-toxic-comment-classification}{Jigsaw Toxic@Kaggle} & Jigsaw Multilingual Toxic Comment Classification & 2020 & toxicity & en, es, tr, pt & 1621 \\\addlinespace[1.5pt]
        \href{https://competitions.codalab.org/competitions/22825}{OSACT4} \cite{mubarak2020overview} & Arabic Offensive Language Detection & 2020 & offense & ar & 27 \\\addlinespace[1.5pt]
        \href{https://hasocfire.github.io/hasoc/2021/index.html}{HASOC@FIRE} \cite{modha2021hasoc} & Offensive Content Identification in English and Indo-Aryan Languages and Conversational Hate Speech & 2021 & hate speech, offense  & en, hi, mr & 65 \\\addlinespace[1.5pt]
        \href{http://nlp.uned.es/exist2021/}{EXIST@IberLEF} \cite{rodriguez2021overview} & Sexism Identification in Social Networks & 2021 & sexism & en, es & 31 \\\addlinespace[1.5pt]
        \href{https://www.urduthreat2021.cicling.org/}{UrduThreat@FIRE} \cite{amjad2022urdu} & Abusive and Threatening Language Detection in Urdu  & 2021 & abuse, threat & ur & 19 \\\addlinespace[1.5pt]
        \href{https://www.kaggle.com/competitions/iiitd-abuse-detection-challenge}{IIIT-D@Kaggle} & Moj Multilingual Abusive Comment Identification across Indic Languages & 2021 & abuse & 10+ Indic & 54 \\\addlinespace[1.5pt]
        \href{https://hasocfire.github.io/hasoc/2022/index.html}{HASOC@FIRE} \cite{satapara2023hasoc} & Offensive Content Identification in English and Indo-Aryan Languages & 2022 & hate speech, offense  & en, hi, de, mr & 12 \\\addlinespace[1.5pt]
        \href{http://nlp.uned.es/exist2022/}{EXIST@IberLEF} \cite{rodriguez2022overview} & Sexism Identification in Social Networks & 2022 & sexism & en, es & 19 \\\addlinespace[1.5pt]
        \href{http://www.di.unito.it/~tutreeb/haspeede-evalita23/index.html}{HaSpeeDe@Evalita} \cite{lai2023haspeede3} & Political and Religious Hate Speech Detection & 2023 & hate speech & it & 6 \\\addlinespace[1.5pt]
        \href{http://nlp.uned.es/exist2023/}{EXIST@IberLEF} \cite{plaza2023overview} & Sexism Identification in Social Networks & 2023 & sexism & en, es & 28 \\\addlinespace[1.5pt]
        \href{https://hasocfire.github.io/hasoc/2023/index.html}{HASOC@FIRE} & Hate Speech and Offensive Content Identifica-  & 2023 & hate speech, offense & en, hi, de, & - \\
        & tion in English and Indo-Aryan Languages &&  & Indo-Aryan &\\
        
        \bottomrule
    \end{tabular}
    \normalsize
    \caption{Summary of competitions and shared tasks in automated identification of cross-lingual HS. Relevant monolingual tasks in non-English languages are also included.
    ``\#Teams'' = ``number of teams participated in the competition and submitted runs''. All task links are added to the ``Name'' column.}
    \label{tab:competition}
\end{table*}

\vspace{0.8mm}
\noindent\textbf{Competitions and Shared Tasks.}
The rising concern surrounding online abuse has led to the establishment of numerous competitions and shared tasks within both national (e.g., GermEval, Evalita, IberLEF) and international (e.g., SemEval) evaluation campaigns. 
These open scientific competitions release benchmark datasets and invite participants to submit detection results and detailed system reports \cite{poletto2021resources}. 
Table \ref{tab:competition} lists 24 such competitions and shared tasks from 2018 to 2023, each addressing different facets of offensive content across multiple languages.

Prominent topics include offense and HS. 
Specifically, GermEval competitions in 2018 \cite{wiegand2018overview} and 2019 \cite{strus2019germeval} centre around identifying offensive content in German tweets. 
OffensEval, initially in 2019 focused only on English content \cite{zampieri2019semeval}, expanded to include multiple languages in the subsequent edition in 2020 \cite{zampieri2020semeval}. 
The task in OSACT4 \cite{mubarak2020overview}, the 4th Workshop on Open-Source Arabic Corpora and Processing Tools, focuses on offensive language detection in Arabic, especially its dialectal forms. 
HaSpeeDe has consistently addressed HS with various general and particular topics \cite{bosco2018evalita,sanguinetti2020haspeede,lai2023haspeede3}.  
For instance, HaSpeeDe \cite{bosco2018evalita} and HaSpeeDe 2 \cite{sanguinetti2020haspeede} identify HS against immigrants and Muslims, while HaSpeeDe 3 \cite{lai2023haspeede3} explores HS in strongly polarised debates, particularly political and religious topics. 
Similarly, the HASOC series, recurring annually from 2019 to 2023, identify HS and offensive content across various languages \cite{mandl2019overview,mandl2021hasoc,modha2021hasoc,satapara2023hasoc}.
TRAC workshops, held in 2018 \cite{kumar2018aggression} and 2020 \cite{kumar2020evaluating}, tackled trolling, aggression, and cyberbullying. 
Other competitions, like PolEval \cite{ptaszynski2019poleval} and UrduThreat \cite{amjad2022urdu}, focus on specific languages (such as Polish and Urdu) or topics (e.g., cyberbullying and threatening languages).
The Jigsaw Multilingual Toxic Comment Classification competition on Kaggle attracted over 1600 participants to build multilingual models using English-only training data.
Another Kaggle competition, IIIT-D Multilingual Abusive Comment Identification, focuses on abusive comments in Indic languages. 
Additionally, some tasks focus on more specific topics, such as sexism and misogyny.
AMI is organised by IberEval \cite{fersini2018ibereval} and Evalita \cite{fersini2018evalita} in 2018, specifically addressing Spanish and Italian misogynous content respectively.
EXIST is held for three consecutive years from 2021 to 2023 and aims to identify sexism in social networks \cite{rodriguez2021overview,rodriguez2022overview,plaza2023overview}. 
Most of these benchmark datasets originate from social media platforms, and often conduct annotations that go beyond simple binary labels to finer-grained ones.



\section{Cross-lingual Resources}
\label{sec-cl-resource}

This section reviews the cross-lingual resources that enable offensive language detection, including datasets, lexicons, models, and auxiliary tools that facilitate transfer across languages.




\vspace{0.8mm}
\noindent\textbf{Multilingual linguistic Resources.}
There are some basic but essential linguistic resources that facilitate the task of cross-lingual HS detection across two or more languages.
Among these, multilingual lexicons and parallel corpora stand out as foundational resources, frequently utilised by researchers to bridge linguistic gaps and enhance model performance in CLTL.

\begin{itemize}
    \item \textbf{Multilingual Lexicons (word-aligned):} contain words, phrases, or terms in two or more languages. These lexicons often provide direct translations or equivalents of terms across the languages they cover, such as HurtLex.\footnote{https://github.com/valeriobasile/hurtlex}

    \item  \textbf{Parallel Corpora (sentence-aligned):} refer to datasets that consist of texts in two or more languages, where each text in one language has a direct translation in the other languages.
\end{itemize}

Due to the lack of sufficient labelled datasets for HS, multilingual lexicons can help bridge this gap by providing connections between resource-rich and resource-poor languages \cite{pamungkas2021joint,jiang2021cross}, while parallel corpora provide translations of labelled instances from resource-rich to resource-poor languages, to facilitate CLTL \cite{rivaling2021llod,stamou2022cleansing,bigoulaeva2023label}. Cross-lingual word embeddings are also created by utilising these multilingual resources. 
These resources are most often created for specific domains, such as HS and abusive language.
While HS patterns can be universal, they can also be specific to certain cultures or languages. They provide insights into how HS manifests differently across languages and cultures, furthering the understanding of cultural nuances \cite{nozza2021exposing}.

\vspace{0.8mm}
\noindent\textbf{Machine Translation and Transliteration Tools.}
Translation refers to converting content from one language into another while preserving meaning of the original text, whereas transliteration is the process of converting text from one script into another by preserving phonetic form rather than semantic content \cite{hande2021offensive,mathur2018detecting}.
Machine translation tools can translate datasets from one language to another, such as Google Translate,\footnote{https://translate.google.com} Microsoft Translator,\footnote{https://translator.microsoft.com/} DeepL\footnote{https://www.deepl.com/en/translator}) and machine translation models (e.g., mBART \cite{liu2020multilingual} and mT5 \cite{xue2021mt5}). 
Besides, machine transliteration tools are essential for code-mixed datasets \cite{kumar2022multi,mathur2018detecting,hande2021offensive}, such as Google Transliteration,\footnote{https://www.google.co.in/inputtools/try/} Microsoft Transliteration,\footnote{https://www.microsoft.com/en-us/translator/business/translator-api/} AI4Bharat transliteration application,\footnote{https://github.com/AI4Bharat/IndianNLP-Transliteration} and transliteration python packages (e.g., indic-transliteration\footnote{https://github.com/indic-transliteration/indic\_transliteration\_py}).
However, in offensive language detection, some translation tools may lead to translation errors or cultural nuances that are inconsistent with the original meaning \cite{pamungkas2021joint}.
For instance, derogatory terms or slurs without direct equivalents in other languages, or their severity might differ across cultures.
Additionally, idiomatic expressions that convey abuse in one language might lose their offensive meaning when translated literally.
Therefore, translation quality is crucial to maintain the effectiveness and robustness of cross-lingual models.

\vspace{0.8mm}
\noindent\textbf{Multilingual Representations.}
Multilingual representations, as language independent representations, are often used in cross-lingual transfer learning. With these representations we can directly address the difference between source and target languages by projecting the text into a shared feature space. Then it is easier to project the behaviour of the model. In general, we can distinguish between using word-level representations and sentence-level representations. 

Multilingual distributional representations encode words from multiple languages in a single distributional word vector space. Cross-lingual transfer of word embeddings aims to establish the semantic mappings among words in different languages by learning the transformation functions over the corresponding word embedding spaces. In this space, semantically similar words (e.g. `cat' in English and `katze' in German) are close together independently of the language.
Some prominent distributional embeddings are Multilingual GloVe \cite{ferreira2016jointly}, Multilingual FastText  \cite{grave2018learning}, Babylon \cite{smith2017offline}, and Multilingual Unsupervised or Supervised word Embeddings (MUSE) \cite{lample2018word,lample2018unsupervised}.
During the training, multilingual distributional representations usually require additional cross-lingual resources, e.g., bilingual dictionaries or parallel corpora. 


Multilingual contextualised representations on sentence level work on a similar principle, but they use sentences instead of words for alignments. They are usually based on an auto-encoder architecture, in which the model is pushed to create similar and context-aware representations for parallel sentences, such as Language Agnostic Sentence Representations (LASER) \cite{artetxe2019massively} and Language Agnostic BERT Sentence Embeddings (LabSE) \cite{feng2022language}.


\vspace{0.8mm}
\noindent\textbf{Monolingual and Multilingual Pre-trained Language Models.}
Pre-trained language models (PLMs) are a SOTA technique in NLP, as a language model trained on large amounts of data. PLMs can be initialised and further trained or fine-tuned with target data for cross-lingual transfer learning. The best-known examples of PLMs are Embeddings from Language Models (ELMo) \cite{peters2018deep} and Bidirectional Encoder Representations from Transformers (BERT) \citep{devlin2019bert}.
\textbf{Multilingual PLMs} are an extension of monolingual PLMs, trained on multiple languages.
They learn the connections between languages by developing a shared linguistic representation.
There is no need to provide multilingual PLMs with any additional information about interlanguage relations, and cross-lingual transfer is possible between any languages \cite{pikuliak2021cross}.
Prominent models include Multilingual BERT (mBERT)  \cite{devlin2019bert} and XLM-RoBERTa (XLM-R) \cite{conneau2020unsupervised}, demonstrating remarkable capabilities in understanding and processing multiple languages simultaneously.

\vspace{0.8mm}
\noindent\textbf{Language-Agnostic Resources}
Language-agnostic resources (e.g., emojis, capitals, and punctuations) can be seen as common traits among different languages, sharing similar knowledge to enhance the semantic connection across languages.
For instance, emojis are generally associated with emotional expressions, which in turn are associated with various forms of online hate \cite{corazza2020hybrid}.

\section{Cross-lingual Transfer Approaches}
\label{sec-cl-method}

\begin{figure}[h]
  \centering
  \includegraphics[width=0.85\linewidth]{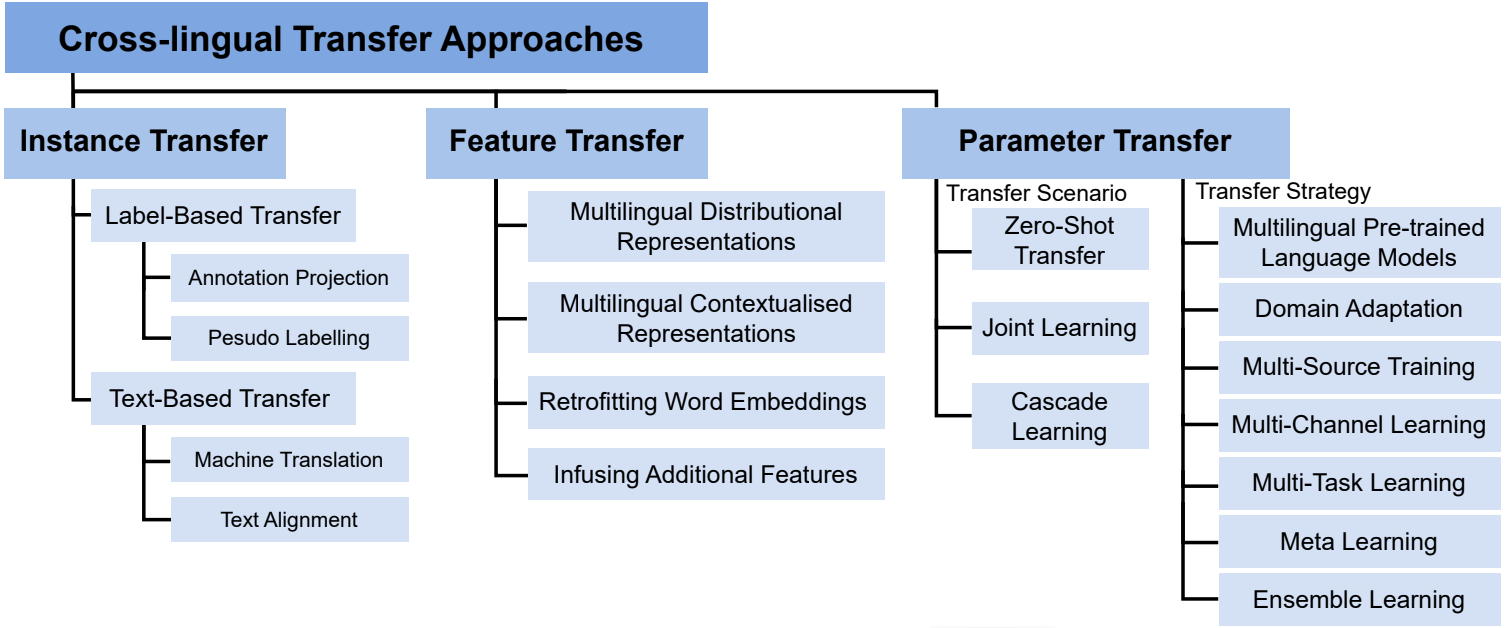}
  \caption{Hierarchy of cross-lingual transfer approaches.}
  \label{fig:cl_method_arch}
\end{figure}

In this section, we systematically describe and compare different cross-lingual techniques for offensive language.
Referring to the way of identifying different transfer learning techniques in \citet{pan2010survey} and \citet{pikuliak2021cross}, we analyse cross-lingual approaches in all reviewed papers and ultimately decide to categorise them according to \textit{``what to transfer''} -- the level at which knowledge transfer or sharing occurs. 
The overall hierarchy of various cross-lingual transfer approaches is depicted in Figure \ref{fig:cl_method_arch}.
Cross-lingual transfer paradigms consist of three main categories:

\begin{itemize}
    \item \textbf{Instance Transfer:} Instances are transferred on the data level between source and target languages, including texts or labels transfer.
    \item \textbf{Feature Transfer:} Linguistic knowledge is shared or transferred on the feature level between source and target languages. 
    \item \textbf{Parameter Transfer:} Parameter values are transferred between language models. This effectively transfers the behaviour of the model from source to target languages.
\end{itemize}

We summarise studies based on transfer levels in Table \ref{tab:transfer-level} and model types in Table \ref{tab:model-type}.
Additionally, a comprehensive Table is provided online\footnote{\url{https://github.com/aggiejiang/crosslingual-offensive-language-survey}} to summarise all techniques utilised in these studies.

\subsection{Instance Transfer}

Instances in the cross-lingual HS task are comprised of two elements (texts and corresponding labels) in both source and target languages.
Although the single source language data is not directly applicable in cross-lingual transfer, there are certain parts of the instance that can still be repurposed together with target language data via instance transfer.
Instance transfer (or instance projection) revolves around the transfer of specific data elements (either text or label information) between source and target languages, which is a key technique in the realm of cross-lingual HS detection.

\begin{table}[h]
    \centering
    \fontsize{6.7}{7.5}\selectfont
    \begin{tabular}{cp{11cm}}
        \toprule
        \textbf{Transfer Level} & \textbf{Reference} \\
        \midrule
        Instance  & \cite{mathur2018detecting,ibrohim2019translated,chiril2019multilingual,pamungkas2019cross,pant2020towards,arango2020hate,elouali2020hate,pamungkas2020misogyny,aluru2020deep,jiang2021cross,song2021study,hande2021offensive,vashistha2021online,pamungkas2021joint,aluru2021deep,salaam2022offensive,ryzhova2022training,boulouard2022detecting,shi2022cross,deshpande2022highly,zia2022improving,das2022data,elalami2022multilingual,kumar2022multi,zhou2023cross} \\
        \addlinespace[2pt]
        Feature & \cite{sohn2019mc,corazza2020hybrid,bansal2020code,aluru2020deep,arango2021cross,hahn2021modeling,rodriguez2021detecting,vitiugin2021efficient,pelicon2021zero,bigoulaeva2021cross,bigoulaeva2022addressing,sarracen2022unsupervised,shi2022cross,jiang2023sexwes,bigoulaeva2023label,paul2023covid} \\
        \addlinespace[2pt]
        Parameter &  \cite{mathur2018detecting,chiril2019multilingual,kapoor2019mind,sohn2019mc,pamungkas2019cross,ousidhoum2019multilingual,pant2020towards,arango2020hate,elouali2020hate,stappen2020cross,glavas2020xhate,pamungkas2020misogyny,ranasinghe2020multilingual,aluru2020deep,markov2021exploring,burtenshaw2021dutch,firmino2021using,ticta2021cross,jiang2021cross,rodriguez2021detecting,song2021study,pelicon2021investigating,pelicon2021zero,bigoulaeva2021cross,ranasinghe2021evaluation,hande2021benchmarking,hande2021offensive,ranasinghe2021mudes,vashistha2021online,ranasinghe2021multilingual,nozza2021exposing,gaikwad2021cross,pamungkas2021joint,aluru2021deep,salaam2022offensive,husain2022tranfer,gupta2022multilingual,montariol2022multilingual,ryzhova2022training,muti2022checkpoint,phung2022exploratory,bigoulaeva2022addressing,boulouard2022detecting,rottger2022data,sarracen2022unsupervised,shi2022cross,deshpande2022highly,eronen2022transfer,mozafari2022cross,zia2022improving,rottger2022multilingual,das2022data,elalami2022multilingual,kumar2022multi,plaza2023respectful,zhou2023cross,bigoulaeva2023label,awal2023model} \\ 
        \bottomrule
    \end{tabular}
    \normalsize 
    \caption{Transfer levels applied in studied papers.}
    \label{tab:transfer-level}
\end{table}

\begin{table}[h]
    \centering
    \fontsize{6.7}{7.5}\selectfont
    \begin{tabular}{p{1.5cm}cp{9.7cm}}
        \toprule
        \textbf{Model Type} & \textbf{Model} & \textbf{Reference} \\
        \midrule
        Machine  & NB & \cite{ibrohim2019translated,burtenshaw2021dutch,rodriguez2021detecting,arango2020hate} \\ \addlinespace[1.5pt]
        Learning & LR & \cite{aluru2020deep,vashistha2021online,aluru2021deep,arango2022resources,deshpande2022highly,paul2023covid} \\\addlinespace[1.5pt]
         & SVM & \cite{ibrohim2019translated,pamungkas2019cross,pamungkas2020misogyny,jiang2021cross,rodriguez2021detecting,bigoulaeva2021cross,paul2023covid} \\\addlinespace[1.5pt]
         & other & \cite{hahn2021modeling} \\ 
         \midrule
        Deep & MLP & \cite{jiang2021cross,pelicon2021zero,kumar2022multi,paul2023covid} \\\addlinespace[1.5pt]
        Learning & CNN & \cite{mathur2018detecting,elouali2020hate,markov2021exploring,jiang2021cross,bigoulaeva2021cross,bigoulaeva2022addressing,jiang2023sexwes,bigoulaeva2023label,paul2023covid} \\\addlinespace[1.5pt]
         & LSTM & \cite{chiril2019multilingual,kapoor2019mind,pamungkas2019cross,ousidhoum2019multilingual,arango2020hate,stappen2020cross,pamungkas2020misogyny,markov2021exploring,burtenshaw2021dutch,jiang2021cross,vitiugin2021efficient,bigoulaeva2021cross,pamungkas2021joint,phung2022exploratory,bigoulaeva2022addressing,shi2022cross,arango2022resources,kumar2022multi,bigoulaeva2023label,paul2023covid}  \\\addlinespace[1.5pt]
         & other & \cite{ousidhoum2019multilingual,bansal2020code,jiang2021cross,vashistha2021online,arango2022resources,deshpande2022highly,kumar2022multi,sarracen2022unsupervised} \\ 
         \midrule
        Transformers & BERT variants & \cite{sohn2019mc,pamungkas2020misogyny,aluru2020deep,markov2021exploring,arango2021cross,hande2021benchmarking,vashistha2021online,muti2022checkpoint,boulouard2022detecting,arango2022resources,zia2022improving,elalami2022multilingual,plaza2023respectful,zhou2023cross,jiang2023sexwes,paul2023covid,husain2022tranfer,pant2020towards} \\ \addlinespace[1.5pt]
         & mBERT & \cite{sohn2019mc,glavas2020xhate,aluru2020deep,burtenshaw2021dutch,ticta2021cross,jiang2021cross,rodriguez2021detecting,song2021study,pelicon2021investigating,pelicon2021zero,hande2021offensive,ranasinghe2021multilingual,nozza2021exposing,pamungkas2021joint,aluru2021deep,montariol2022multilingual,muti2022checkpoint,bigoulaeva2022addressing,boulouard2022detecting,sarracen2022unsupervised,deshpande2022highly,eronen2022transfer,das2022data,elalami2022multilingual,zhou2023cross,bigoulaeva2023label,awal2023model} \\\addlinespace[1.5pt]
         & XLM variants & \cite{corazza2020hybrid,stappen2020cross,hande2021benchmarking,zhou2023cross,montariol2022multilingual,rottger2022data,rottger2022multilingual} \\\addlinespace[1.5pt]
         & XLM-R & \cite{pant2020towards,glavas2020xhate,ranasinghe2020multilingual,firmino2021using,ticta2021cross,jiang2021cross,song2021study,ranasinghe2021evaluation,hande2021benchmarking,hande2021offensive,ranasinghe2021mudes,ranasinghe2021multilingual,nozza2021exposing,gaikwad2021cross,salaam2022offensive,gupta2022multilingual,montariol2022multilingual,ryzhova2022training,sarracen2022unsupervised,arango2022resources,eronen2022transfer,mozafari2022cross,zia2022improving,plaza2023respectful,awal2023model} \\\addlinespace[1.5pt]
         & other & \cite{rodriguez2021detecting,pelicon2021investigating,hande2021benchmarking,hande2021offensive,husain2022tranfer,gupta2022multilingual,montariol2022multilingual,plaza2023respectful,jiang2023sexwes,pant2020towards,das2022data} \\
         \midrule
        LLMs & FLAN-T5, mT0 & \cite{plaza2023respectful} \\
        \bottomrule
    \end{tabular}
    \normalsize 
    \caption{Types of models used in studied papers. Multilingual PLMs are kept in Transformers and others are grouped into *variants.}
    \label{tab:model-type}
\end{table}

To implement instance transfer, a correspondence between instances in the source and target languages must be created on the data level. \textbf{Correspondence} refers to the pair of instances with the same meaning of texts or identical labels \cite{pikuliak2021cross}, e.g., parallel data with the same label matching, and translated text retains the same label of its original counterpart. Based on the established correspondence, texts or labels can be transitioned from one language to the other
. Subsequently, these projected texts or labels can be used for further training and fine-tuning stages. 

Pairs of instances in source and target languages that share identical labels are referred to as corresponding instances.
All data-based transfer techniques need corresponding texts/labels or a mechanism to generate them. 
Techniques for instance transfer are introduced distinctly at the text and label levels.
For label-based transfer, labels from a source language are projected onto target-language data (e.g., via annotation projection or pseudo labelling), without transferring the source text. 
In contrast, text-based transfer is to transfer or generate text across languages (e.g., via machine translation or text alignment), preserving the original labels and creating new training instances in the target language. 
These two strategies differ in which component of an instance is projected across languages, but both establish correspondence between source and target data.




\subsubsection{\textbf{Label-Based Transfer}}



\vspace{-1mm}
\paragraph{\textbf{Annotation Projection}}
\label{sec:label-transfer}

It leverages parallel corpora to transfer annotations from a source to a target language. 
Parallel corpora consist of sentence-aligned source-to-target data, ensuring higher quality than machine-generated translations.
Since data is paired between source and target languages, labels in the source language can be directly projected onto the target language. For example, if a sentence in the source language is labelled as hateful, its corresponding text in the target language can be inferred to the same label.
It can effectively create labelled data for the target language without manual annotation, especially when building datasets for languages where annotating from scratch might be challenging due to linguistic complexities or a lack of annotators.

\vspace{-1mm}
\paragraph{\textbf{Pseudo Labelling}}

It is a semi-supervised learning technique, where a model trained on labelled source data is used to make predictions on unlabeled target data \cite{lee2013pseudo}. 
High-confidence predictions, namely predicted labels with high probability scores, are treated as ``pseudo-labels'' for the unlabelled target data.
By using high-confidence predictions from a model trained in the source language, it is able to generate labelled target data in another language. This augmented target data can then be used to train or fine-tune models for offensive language detection in the target language.
To generate pseudo labels, some studies use an ensemble-based approach to generate labels based on majority voting for unlabelled target datasets, and use bootstrapped target dadasets to further fine-tune the trained model with source training samples 
\cite{bigoulaeva2021cross,bigoulaeva2022addressing,bigoulaeva2023label}.
\citet{hande2021offensive} directly use a pre-trained multilingual language model (i.e. XLM-R \cite{conneau2020unsupervised}) to predict the pseudo labels. 
\citet{zia2022improving} firstly fine-tune a pre-trained XLM-R  on gold-labeled source language data, and then use it to create a new psuedo-labelled dataset in the target language.

\subsubsection{\textbf{Text-Based Transfer}}

\vspace{-1mm}
\paragraph{\textbf{Machine Translation}}

In the absence of sufficient annotated parallel data resources, machine translation provides an efficient alternative strategy to achieve text-level transmission.
In general, it leverages translation tools \cite{pamungkas2020misogyny,pamungkas2021joint} or models \cite{glavas2020xhate} to translate labelled datasets between source and target languages, thereby augmenting the training data. Labels for translated data are the same as those in the original labelled dataset.
Translation can be either directed or undirected: (i) target to source \cite{ibrohim2019translated,arango2020hate,boulouard2022detecting,aluru2020deep,aluru2021deep}; (ii) source to target \cite{glavas2020xhate,song2021study,salaam2022offensive}; (iii) translate both source and target to each other \cite{sohn2019mc,pamungkas2019cross,jiang2021cross,elalami2022multilingual,das2022data}.
In addition, back-translation is commonly used as a data augmentation technique in cross-lingual scenarios by using various transformations on the original source data to create new samples for further model training \cite{ouyang2021ernie}.
Back-translation refers to translating a source sentence to a target language and then reverting it back to the original language, aiming to generate paraphrased instances of the original texts.

\vspace{-1mm}
\paragraph{\textbf{Text Alignment}} 

Similar to machine translation, existing mapping techniques between languages are employed to generate labelled data for the low-resource language.
\citet{shi2022cross} begin by using a shared space between two language vectors to identify the most similar sentence in the target language to the labelled data in the source language. The identified sentences are then treated as labelled data for the target language, and assigned the same labels of the source data. 
\citet{ryzhova2022training} augment training datasets by generating attacked source datasets. They apply an adversarial attack algorithm \cite{jin2020bert} to the source sentence by only replacing words that the model considers important, ensuring that the new sentence is semantically similar to the old one. 

\vspace{-1mm}
\subsection{Feature Transfer}

Feature transfer transforms linguistic features to aid in HS detection. 
Rather than translating source texts or projecting source labels, they focus on extracting salient linguistic features from source and target languages, and aligning them into a shared feature space, retaining the essence of the text consistent across languages.
\textbf{Cross-lingual Word Embeddings (CLWEs)}, also known as \textbf{Multilingual Word Embeddings (MWEs)}, are commonly used for feature transfer. They train monolingual word embeddings (WEs), such as Word2Vec or FastText, on multiple languages, yielding vectors that capture semantic similarities relying on shared representations between languages and thus enabling representations that are common across languages.
In cross-lingual offensive language detection, various approaches utilise existing CLWEs or retrofit CLWEs. We discuss three main techniques: multilingual distributional representations, multilingual contextualised representations, and retrofitting WEs.


\vspace{-1mm}
\subsubsection{\textbf{Multilingual Distributional Representations}}

Pre-trained distributional WEs are most commonly used as embedding layers in neural networks for different language data inputs.
Among these, MUSE embeddings have emerged as a predominant choice.
They are extensively utilised to extract multilingual features and are often integrated with both traditional machine learning models, such as Gradient Boosted Decision Trees \cite{arango2020hate}, and  advanced deep learning architectures (i.e. LSTM \cite{pamungkas2019cross,pamungkas2020misogyny,pamungkas2021joint,arango2021cross,burtenshaw2021dutch}, 
BiLSTM \cite{bigoulaeva2021cross,bigoulaeva2022addressing,bigoulaeva2023label,jiang2021cross}, 
CNN-GRU \cite{deshpande2022highly,aluru2021deep,aluru2020deep}, 
Capsule Networks \cite{jiang2021cross}, Sluice networks \cite{ousidhoum2019multilingual}), and BERT \cite{pamungkas2020misogyny}).
Other notable multilingual distributional vectors, such as Multilingual GloVe and FastText, have also been employed in architectures like BiLSTM \cite{chiril2019multilingual} and Capsule Networks \cite{jiang2021cross}.
In addition, Babylon multilingual embeddings have been paired with Sluice networks for multilingual HS detection \cite{ousidhoum2019multilingual}.
\citet{bansal2020code} integrate bilingual switching features into the Hierarchical Attention Network (HAN) architecture, enhancing the transfer knowledge for cross-lingual detection.

Furthermore, some studies have ventured into cross-lingual projection using the MUSE mapping method, aligning monolingual FastText embeddings in English and German to produce their own CLWEs \cite{bigoulaeva2021cross,bigoulaeva2022addressing,bigoulaeva2023label}.
\citet{paul2023covid} align English and Hindi FastText embeddings by performing Canonical Correlation Analysis (CCA),\footnote{https://www.mathworks.com/help/stats/canoncorr.html} and project them into a shared vector space where they are maximally correlated.
\citet{kapoor2019mind} use multilingual datasets to train embeddings in order to capture specific distributional representations of tweets.
\citet{sarracen2022unsupervised} propose a graph auto-encoder framework to learn embeddings of a set of texts in an unsupervised way, adding language-specific knowledge via Universal Sentence Encoder (USE).


\vspace{-1mm}
\subsubsection{\textbf{Multilingual Contextualised Representations}}

Multilingual contextualised representations, typically derived at the sentence level from PLMs (such as mBERT \cite{devlin2019bert} and XLM-R \cite{conneau2020unsupervised}), can capture deeper semantic and contextual relationships between words and phrases. 
Some researchers have leveraged hidden features with richer semantic information from the first embedding layer of mBERT \cite{arango2021cross, pamungkas2021joint,rodriguez2021detecting}, DistilmBERT \cite{vitiugin2021efficient,hande2021offensive}, XLM \cite{stappen2020cross}, XLM-R \cite{hande2021offensive} as feature extraction layers. These embeddings are either extracted as standalone embedding layers \cite{arango2021cross, pamungkas2021joint,vitiugin2021efficient,rodriguez2021detecting} or utilised as frozen embedding layers \cite{stappen2020cross,hande2021offensive} in deep neural networks, such as BiLSTM, yielding enhanced performance over traditional distributional embeddings. 
Multilingual sentence embeddings LASER have also been used across various models, including Support Vector Machine \cite{rodriguez2021detecting}, Logistic Regression \cite{aluru2021deep,aluru2020deep,pamungkas2021joint,deshpande2022highly}, Random Forest \cite{arango2021cross,rodriguez2021detecting}, XGBoost \cite{arango2021cross}, Multi-Layer Perceptron \cite{pelicon2021zero} and LSTM \cite{vitiugin2021efficient}.
\citet{rodriguez2021detecting} propose to use LabSE and mBERT representations with SVM-based and tree-based classifiers, achieving improved performance.

\vspace{-1mm}
\subsubsection{\textbf{Retrofitting Word Embeddings}}

Retrofitting pre-trained WEs by infusing multilingual domain knowledge stands as a promising strategy to amplify the semantic relationships across languages, thereby enhancing the efficacy and scalability of models in cross-lingual HS detection. 
\citet{pant2020towards} employ a supervised FastText model trained on the sarcasm detection corpus \cite{swami2018corpus} to improve the identification of HS content in English and Hindi languages.
\citet{arango2021cross} construct HS-specific WEs by aligning monolingual Word2Vec embeddings using hate-specific bilingual dictionaries (such as HurtLex), which are able to capture non-traditional translations of words between languages.
Additionally, \citet{hahn2021modeling} learn semantic subspace-based representations to model profane languages on both word and sentence levels. \citet{jiang2023sexwes} propose a cross-lingual domain-aware semantic specialisation system to construct sexism-aware WEs. They retrofit pre-trained FastText word vectors by integrating in-domain and out-of-domain linguistic knowledge (such as lexico-semantic relations) into the specialised feature space.

\vspace{-1mm}
\subsubsection{\textbf{Infusing Additional Features}}

Additional external features are commonly used in identifying HS across languages, such as domain-specific \cite{pamungkas2019cross,hahn2021modeling}, language-specific \cite{zhou2023cross,arango2021cross}, and typographic features \cite{kumar2022multi,markov2021exploring}.
Specifically, domain-specific features are essential to assist the hate-related knowledge transfer. 
The multilingual lexicon, HurtLex, has been employed in multiple studies \cite{pamungkas2020misogyny, pamungkas2021joint, pamungkas2019cross, jiang2021cross}. 
Some researchers also utilise hateful word pairs to construct domain-aware or semantic-based representations across languages \cite{jiang2023sexwes,hahn2021modeling,arango2021cross}, due to the lack of hate-specific patterns in general-purpose multilingual embeddings.
In addition, infusing language-specific features can help integrate linguistic and cultural knowledge for geographically sensitive tasks.
It can be in different forms, such as language switching pattern matrix \cite{bansal2020code}, bilingual pairs \cite{jiang2023sexwes}, cross-cultural similarities \cite{arango2021cross}, and social dynamics among users \cite{nozza2021exposing}.
This feature can provide a richer understanding of cultural dimensions in hateful content \cite{zhou2023cross}. 
Furthermore, typographic features provide a language-agnostic and domain-agnostic perspective to help identify online hate due to their shared meanings across languages, which includes capitals, punctuations, and emojis  \cite{kumar2022multi,markov2021exploring,corazza2020hybrid}. Stylometric features also remain persistent with respect to language variations, as the writing style of toxic content can be correlated to emotional profile of social media users \cite{markov2021exploring}.

\vspace{-1mm}
\subsection{Parameter Transfer}

\begin{figure}[ht]
  \centering
  \includegraphics[width=0.75\linewidth]{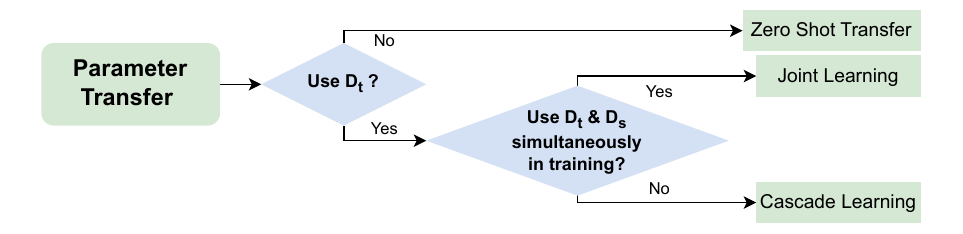}
  \caption{Different scenarios in parameter transfer for automated detection of cross-lingual HS.}
  \label{fig:scenario}
\end{figure}

The performance of NLP models is dependent on model parameters.
Parameter transfer assumes that prior distributions of parameters are transferred between different languages within one model or between individual models. Most parameter transfer approaches rely on inductive transfer learning in cross-lingual offensive language detection \cite{pan2010survey}.
Inspired by \citet{pikuliak2021cross}, we outline three distinct scenarios for surveyed studies, depending on how the source and target data are utilised during the process of parameter transfer, namely zero-shot transfer, joint learning, and cascade learning (see Figure \ref{fig:scenario}).
Only source data is used for training in zero-shot transfer, while both source and target data are applied during the training process in joint learning and cascade learning. Joint learning utilises both simultaneously to train the model, but cascade learning leverages them in the different stages of training.

Parameter-level transfer strategies can be sophisticated and diverse in terms of their model architectures and training procedures, incorporating one or more of the transfer scenarios above.
Table \ref{tab:param-transfer} has summarised the possible correlations between these transfer strategies and their application scenarios, and will discuss more details in the following subsections regarding scenarios and strategies utilised on parameter-level transfer.

\begin{table*}[ht!]
    \centering
    \fontsize{6.7}{7.5}\selectfont
    \begin{tabular}{p{3.3cm}ccc}
        \toprule
        \textbf{Transfer Strategy} & \textbf{Zero-Shot Transfer} & \textbf{Joint Learning} & \textbf{Cascade Learning} \\
        \midrule 
        Multilingual PLMs & \cmark\cmark & \cmark & \cmark\cmark \\\addlinespace[1.5pt]
        Domain Adaptation & & \cmark & \cmark\cmark \\\addlinespace[1.5pt]
        Multi-source Training & \cmark\cmark & \cmark\cmark & \cmark \\\addlinespace[1.5pt]
        Multi-channel Learning & \cmark & \cmark\cmark & \\\addlinespace[1.5pt]
        Multi-task Learning & \cmark & \cmark\cmark & \cmark \\\addlinespace[1.5pt]
        Meta Learning & & \cmark & \cmark \\\addlinespace[1.5pt]
        Ensemble Learning & \cmark & \cmark\cmark & \cmark \\
        \bottomrule 
    \end{tabular}
    \normalsize
    \caption{Correlations between scenarios and strategies in parameter transfer for automated detection of cross-lingual HS. \cmark\cmark means higher frequency than \cmark.}
    \label{tab:param-transfer}
\end{table*}

\vspace{-1mm}
\subsubsection{\textbf{Transfer Scenarios}}
\label{sec:trans-scenario}

We discuss three different scenarios of parameter transfer.

\vspace{0.8mm} 
\noindent\textbf{Scenario 1: Zero-Shot Transfer.}
It is a promising scenario when labelled data in the target language is scarce or lacking, by enabling detection in a language not seen during training.
The model is trained on labelled source data from one or more languages not including the target language, and is then tested on the unseen target language. 
This is sometimes called direct or model transfer. 
We found 25 papers under this transfer paradigm, including subtypes:

\begin{itemize}
    \item \textit{Single Source to Single Target:} Considered a type of domain adaptation (see \S\ref{sec:strategy}), models are trained on labelled data from a single source language, then used on an unseen target language \cite{glavas2020xhate,firmino2021using,ticta2021cross,rodriguez2021detecting,pelicon2021zero,nozza2021exposing,montariol2022multilingual,eronen2022transfer,zhou2023cross,gupta2022multilingual,ranasinghe2021evaluation,ranasinghe2021mudes,gaikwad2021cross,muti2022checkpoint,das2022data,plaza2023respectful}. It is the most frequently observed (62.5\%) of zero-shot transfer strategies in our survey.

    \item \textit{Multiple Sources to Single Target:} Models are trained on labelled data from multiple source languages (excluding the target language) and then evaluated on a single target language \cite{ryzhova2022training,montariol2022multilingual,aluru2020deep}. It is a type of multi-source training strategy where the combination of training data excludes the target (see \S\ref{sec:strategy}).

    \item \textit{Pseudo-Target Augmented Training:} The source data is translated into the parallel target data via machine translation, and then models are trained on both the original source data and the translated target data together \cite{stappen2020cross} or in parallel \cite{pamungkas2019cross,pamungkas2021joint,pamungkas2020misogyny,jiang2021cross}.
    It falls under the multi-channel learning strategy when using both source and translated target data in parallel\footnote{Pseudo-Target Augmented Training has overlaps with the joint learning scenario. It distinguishes itself by using pseudo target texts for training instead of real target texts.} and fusing them before the output layer of the model. This allows the model to learn nuances specific to the target language from the pseudo content.
    
    \item \textit{Parameter Frozen:} A model is trained on source data first, and then a subset of its parameters (e.g., weights from specific layers) is saved to initialise another model for target language evaluation \cite{ranasinghe2020multilingual,ranasinghe2021multilingual,gaikwad2021cross,phung2022exploratory}.
    It is a type of domain adaptation,\footnote{Parameter frozen in zero-shot transfer scenario shares similarities with the cascade learning scenario. However, while cascade learning continues to fine-tune the trained model with target data, zero-shot transfer directly applies the trained model to test the target data without further fine-tuning.} and captures language knowledge from certain layers of the source model to provide a head start for the target model, potentially leading to faster convergence and better generalisation.
\end{itemize}

It is worth highlighting that these methods rely on multilingual PLMs, especially often used for Single Source to Single Target and Multiple Sources to Single Target scenarios. mBERT and XLM-R are two of the most prominent multilingual PLMs in surveyed papers, because of their popularity, widespread adoption and SOTA performance across multilingual NLP tasks. They together account for 80.8\% of papers in the zero-shot transfer scenario. Additionally, some papers release domain- and language-specific PLMs in a multilingual setting \cite{gupta2022multilingual}, choose other multilingual PLMs like XLM \cite{stappen2020cross} and language-specific MuRIL \cite{gupta2022multilingual,das2022data}, or combine MWEs with monolingual neural models like LSTM \cite{pamungkas2019cross,jiang2021cross} and BERT \cite{pamungkas2020misogyny}.

\vspace{0.8mm} 
\noindent\textbf{Scenario 2: Joint Learning.}
It typically involves training a model between languages based on multiple channels, tasks or objectives simultaneously, or jointly assimilating multiple types of data or modalities.
The model might have some parameters that are shared across languages as well as some that are specific to each, so it is also called \textit{parameter sharing} \cite{pikuliak2021cross}.
The transfer of knowledge happens only on shared parameters. 
Changes to the shared parameters can affect each other between languages. They are interrelated and can provide complementary information.

\begin{figure}[h]
  \centering
  \includegraphics[width=0.8\linewidth]{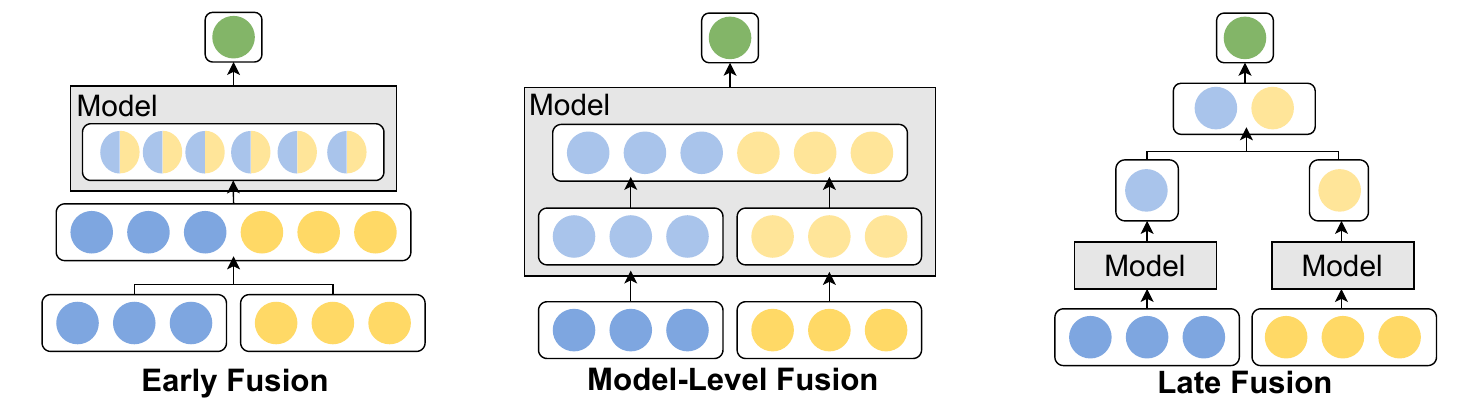}
  \caption{Different fusion stages in the joint learning scenario.}
  \label{fig:fusion}
\end{figure}

In this scenario, both source and target data are used at the same time during the training stage.
Parameters can be shared between source and target channels within one model or between individual source and target models.
We found 23 papers in this group.
The distinction among these strategies mainly lies in the stage of knowledge transfer across languages and the extent of parameter sharing (some or all parameters shared) \cite{kumar2022multi}. Based on these criteria, we categorise them into three primary fusion stages (see Figure \ref{fig:fusion}):

\begin{itemize}
    \item \textit{Early fusion:} A combination of datasets or embeddings from both source and target languages, with the entire model sharing parameters. It is commonly adopted in few-shot learning \cite{stappen2020cross,gaikwad2021cross,firmino2021using,das2022data,gupta2022multilingual,zhou2023cross} and multi-source training \cite{muti2022checkpoint,salaam2022offensive,ranasinghe2021evaluation,deshpande2022highly,vashistha2021online,elouali2020hate,das2022data,gupta2022multilingual,markov2021exploring,chiril2019multilingual}, where datasets from the target language are incorporated during training. 
    In few-shot learning, the model is trained by utilising a very small amount of target samples combined with a more extensive source dataset, while multi-source training typically takes a combined dataset of source and target languages or integrates multiple datasets including multiple non-target datasets together with the target dataset for training.
    
    \item \textit{Model-level fusion:} A concatenation of high-level hidden features within a model, and only specific layers or components of the model share parameters, facilitating a more nuanced transfer of knowledge. It commonly occurs in parallel model architectures, such as multi-channel learning \cite{sohn2019mc,kumar2022multi}, multi-task learning \cite{hande2021benchmarking,ousidhoum2019multilingual} and meta-learning \cite{mozafari2022cross,awal2023model}.

    \item \textit{Late fusion:} A combination of different channels or models before the output layer, and only the parameters of the output layer for predictions are shared, ensuring a final consolidation of knowledge from various languages, often used in multi-channel architectures  \cite{ibrohim2019translated}.
\end{itemize}

Early fusion is the most prevalent strategy due to its straightforward integration of source and target data for better generalisation, while late fusion provides a modular approach, allowing for the independent training of individual models that can later be merged.
However, unlike early fusion, model-level fusion effectively overcomes the curse of high dimensionality and the synchronisation demands among diverse features. Moreover, it ensures that interactions between different languages are not isolated as in late fusion \cite{kumar2022multi}, showing its robustness and resilience.

\vspace{0.8mm} 
\noindent\textbf{Scenario 3: Cascade Learning.}
In the cascade learning scenario, both source and target data are used at different times during the training. First, an existing PLM is directly used or a new model is trained with source data, and then the trained model is fine-tuned with target data, where subsets of parameters are shared between different stages of training and fine-tuning.
Different strategies for training and fine-tuning have been used in 19 of our surveyed papers.



Many use existing SOTA pre-trained multilingual models serving as the foundation,\footnote{Cascade learning fine-tunes PLMs using target data, while zero-shot transfer fine-tunes PLMs using non-target data.} while some studies train a model with one or multiple source datasets.
Subsequently, the fine-tuning stage in cascade learning is applied across an entire \cite{kapoor2019mind,firmino2021using,aluru2021deep,pelicon2021investigating,husain2022tranfer} or only a limited subset of target data (i.e. few-shot learning) \cite{ranasinghe2021evaluation,pelicon2021investigating}, optimising the model adaptability.
\citet{rottger2022data} indicate that an initial training phase on source data (English) could increase model performance when there is little fine-tuning data in the target language, where source data can partly substitute target data in few-shot settings. 
Some researchers delve into consistent fine-tuning \cite{ryzhova2022training,mozafari2022cross,awal2023model} or iterative fine-tuning (i.e. meta learning) \cite{mozafari2022cross,awal2023model}, performing multiple rounds of fine-tuning on the PLMs to further enhance its ability to transfer knowledge.

Additionally, given the limited availability of target data, data augmentation techniques have emerged. They aim to enrich the target dataset to enable fine-tuning, by enhancing label diversity or expanding the text corpus. Some studies have attempted to generate pseudo labels for unlabelled target datasets \cite{bigoulaeva2021cross,bigoulaeva2022addressing,bigoulaeva2023label,zia2022improving} (see \S\ref{sec:label-transfer}), while others augment and refine pseudo texts by exploring advanced techniques, such as word alignment to project samples \cite{shi2022cross}, adversarial algorithm to generate attack samples \cite{ryzhova2022training}, and domain-specific target data filtering \cite{glavas2020xhate}.

\vspace{-1mm}
\subsubsection{\textbf{Hybrid Transfer Strategies}}
\label{sec:strategy}

According to the utilisation of source and target data, these strategies can be broadly covered in one or more parameter transfer scenarios mentioned in \S\ref{sec:trans-scenario}.
That is, these strategies are not mutually exclusive. A single scenario can seamlessly integrate multiple transfer strategies (see Table \ref{tab:param-transfer}).

\vspace{0.8mm} 
\noindent\textbf{Multilingual Pre-Trained Language Models.}
These models are regarded as a cornerstone of parameter-level transfer techniques for low-resource scenarios, especially in cross-lingual offensive language detection. 
They are trained on large-scale multilingual corpora, learning commonalities and differences between languages. Therefore, a single PLM can share the same underlying sub-word vocabulary and semantic representations across languages \cite{devlin2019bert}.
Multilingual PLMs can be initialised as the base model with pre-trained parameters, then fine-tuned on available HS data in all three scenarios of parameter transfer (zero-shot transfer, joint learning and cascade learning) \cite{muti2022checkpoint,firmino2021using}.
Beyond this, they can also employed in the other two transfer levels (data and feature). 
They are able to augment target datasets by predicting pseudo labels for unlabelled data and generating synthetic target texts. 
And they can be used as multilingual representations, where subsequent models are constructed based on these source representations \cite{rodriguez2021detecting}.






\vspace{0.8mm} 
\noindent\textbf{Domain Adaptation.}
It involves taking a model trained (or pre-trained) in the source language and adapting its parameters to initialise a new model for the low-resource language, aiming to leverage source data to improve performance on target data \cite{lee2021exploring,glavas2020xhate}.
Its advantage is that it can capture offensive language knowledge and linguistic similarities between the languages within the model's parameters.
In the zero-shot transfer scenario, parameters of the trained model are entirely or partially used on the target data for prediction \cite{gaikwad2021cross,ranasinghe2021multilingual}.
In addition, cascade learning adopts a more subtle approach. That is, the entire or specific subsets of the model's parameters (like specific layers) are frozen to retain their original state \cite{ranasinghe2021evaluation}, while others are fine-tuned to better align with the target data, often using a smaller target dataset (as few-shot learning) \cite{pelicon2021investigating,rottger2022data}. 

\vspace{0.8mm} 
\noindent\textbf{Multi-Source Training.}
This approach often entails the amalgamation of labelled data from multiple source languages and potentially the target language in diverse ways.
By synthesising information across multiple languages, this strategy is able to capture the diversities and commonalities of linguistic patterns, demonstrating significant efficacy in the cross-lingual setting. 
It is extensively utilised across all three transfer scenarios, serving as an augmented cross-lingual learning approach during different stages of training or fine-tuning.
In multi-source training, the most prevalent way is to train or fine-tune the model on multiple source languages. This can be executed in the training phase of joint learning \cite{rottger2022multilingual,muti2022checkpoint,salaam2022offensive,deshpande2022highly,vashistha2021online,elouali2020hate,elalami2022multilingual,das2022data,hande2021benchmarking,markov2021exploring,chiril2019multilingual} or the fine-tuning phase of cascade learning \cite{ranasinghe2021evaluation,gupta2022multilingual} in conjunction with the target language, or in isolation from the target language for zero-shot transfer \cite{ryzhova2022training,montariol2022multilingual}.
\citet{deshpande2022highly} offer a more focused perspective by narrowing the linguistic scope to specific language families. By training models on languages within one language family, they then assess the model's performance on each member of that family, revealing the interplay of linguistic knowledge within related languages. 
It is worth noting that merging training examples from different languages can be detrimental to cross-lingual performance in cases where offensive language domains are too distant \cite{glavas2020xhate}.




\vspace{0.8mm} 
\noindent\textbf{Multi-Channel Learning.}
This strategy typically refers to the use of multiple types of input representations or sources concurrently for a single task, where multiple embeddings or pre-processed versions of the text are often channelled as parallel input streams.
In this architecture, each channel processes the input independently through parallel layers.
These independent processing channels are eventually combined based on their outputs for subsequent learning or classification stages. Such a design ensures that diverse input representations are holistically integrated, capturing a broader spectrum of linguistic knowledge across languages.
Multi-channel learning finds its applications mainly in the zero-shot transfer and joint learning scenarios. 
When applied to cross-lingual offensive language detection, this approach often treats datasets from different languages as distinct input channels. 
Each channel adopts different neural networks or PLMs based on diverse input languages, and these channels are then merged into the final classification stage \cite{kumar2022multi,stappen2020cross,sohn2019mc}. 
A significant improvement to this strategy is the incorporation of machine translation. 
By translating labelled source data to pseudo target data, low-resource target data is enriched and a bridge is constructed between source and target languages, enhancing knowledge relationships and a more seamless transfer of languages \cite{pamungkas2019cross,pamungkas2020misogyny,pamungkas2021joint,jiang2021cross,ibrohim2019translated}.





\vspace{0.8mm} 
\noindent\textbf{Multi-Task Learning.}
A joint learning environment training a model on several interrelated tasks simultaneously, where the proficiency learned from one task can boost performance on another.
A key point of this strategy is the shared representation of hidden features across specific layers of the model.
Such shared layers enable different tasks to mutually benefit from common representations, while also preserving their distinctiveness through task-specific output layers. 
The training process is holistic, aiming to optimise the model's performance across all tasks. This is typically achieved by incorporating the loss functions from each task.
Multi-task learning emerges as a powerful method for cross-lingual offensive language detection within the parameter-level transfer techniques across three transfer scenarios.
Models can be trained on offensive language detection and a wide range of related auxiliary tasks, such as Sentiment Analysis, Named Entity Recognition (NER), Dependency Parsing, and Part-Of-Speech (POS) Tagging 
\cite{montariol2022multilingual,hande2021benchmarking,ousidhoum2019multilingual}. Additionally, they can be fine-tuned on data in various offensive categories, such as aggressive content \cite{ranasinghe2020multilingual,ranasinghe2021multilingual}. 
This multi-task training strategy enables the model to identify and process offensive content with higher accuracy and enhances the model's ability to grasp linguistic nuances across languages.








\vspace{0.8mm} 
\noindent\textbf{Meta Learning.}
This strategy, often referred to as ``learning to learn'', has become a popular few-shot learning technique in NLP.
It involves training models to learn the optimal initialisation of parameters, allowing them to be fine-tuned with a small amount of data from the target language.
This technique finds its application within the joint learning and cascade learning transfer scenarios.
It enables rapid adaptation to new unseen languages, making it especially useful in scenarios where labelled data is limited.
The main meta learning methods include optimisation-based \cite{finn2017model} and metric-based techniques \cite{koch2015siamese}.
Two papers in our review employ optimisation-based meta learning framework.
\citet{montariol2022multilingual} study meta learning for the problem of few-shot HS detection in low-resource languages. They propose a cross-lingual meta learning-based approach based on optimisation-based Model-Agnostic Meta-Learning (MAML) and Proto-MAML models, fine-tuning the base learner XLM-R with parallel few-shot datasets in different target languages.
\citet{awal2023model} propose HateMAML, a model-agnostic meta-learning-based framework that uses a semi-supervised self-refinement strategy to fine-tune a better PLM for unseen data in target language, showing effective performance for HS detection in low-resource languages.






\vspace{0.8mm} 
\noindent\textbf{Ensemble Learning.}
It leverages multiple models or ``learners'' to collaboratively make decisions. Instead of relying on the output of a single model, ensemble learning combines all predictions from multiple models to produce a final prediction based on majority voting \cite{paul2023covid,markov2021exploring}.
It takes advantage of the different strengths of each model, mitigating the weaknesses and biases of any single model and avoiding overfitting, improving robustness and reliability over a single classifier.
This is particularly beneficial in cross-lingual settings, where language biases can pose challenges to individual models.
In cross-lingual offensive language detection, ensemble learning is mainly applied in the zero-shot, joint learning, and cascade learning transfer scenarios. 
Deep ensemble models have emerged as the predominant choice in recent studies \cite{bigoulaeva2021cross,bigoulaeva2022addressing,bigoulaeva2023label,paul2023covid}, while the integration of PLMs within ensemble learning frameworks is also common \cite{burtenshaw2021dutch,ryzhova2022training,markov2021exploring}.
Each model in the ensemble can be trained on different languages or linguistic features, allowing the ensemble model to capture diverse linguistic patterns and reduce test errors for target languages \cite{ryzhova2022training}.

\vspace{-1mm}
\subsection{Summary of Cross-lingual Approaches}

A review of the surveyed work reveals several key developments in cross-lingual offensive language detection.
First, there has been a \textbf{rise of multilingual PLMs}, beginning with general-purpose PLMs such as mBERT and XLM-R, which quickly replaced feature- or translation-based methods as the dominant approach~\cite{devlin2019bert,conneau2020unsupervised}. Meanwhile, there is a move \textbf{from general-purpose PLMs to more language- and culture-specific PLMs}, such as MuRIL for Indic languages~\cite{das2022data}, AraBERT for Arabic dialects~\cite{husain2022tranfer}, and XLM-T trained on social media~\cite{montariol2022multilingual}. More recently, a new trend has emerged with the rise of \textbf{large language models (LLMs)}, capable of zero- or few-shot multilingual tasks via prompting~\cite{brown2020language,openai2023gpt4,plaza2023respectful}. See future challenges and directions in PLMs and LLMs in \S\ref{sec:future}.
Second, we observe a \textbf{shift from feature-based to parameter-based transfer}. Earlier approaches relied on CLWEs, lexicons, or projection methods~\cite{pamungkas2019cross,arango2021cross}. With the rise of multilingual PLMs, parameter-based fine-tuning has become the core transfer mechanism, while handcrafted features now play a complementary role.  
Moreover, there is a \textbf{growing focus on low-resource languages and dialects}. While early research centred on English and a few European languages, more recent work increasingly covers Indic~\cite{das2022data}, Arabic~\cite{husain2022tranfer}, and Asian languages~\cite{jiang2022swsr,jeong2022kold}. However, many languages still lack dedicated datasets and evaluation benchmarks.


\vspace{0.8mm} 
\noindent\textbf{Model distributions.} 
After reviewing diverse models and their frequency of use in recent studies, we can see
traditional machine learning-based models have been employed 24 times, while deep learning-based models have seen a slightly higher usage (38 times) due to their complex structure and learning capacity from a large amount of data.
Transformer-based models have been overwhelmingly preferred (99 times), highlighting their growing dominance in NLP.
Large Language Models have also recently been investigated in this area due to their excellent emergent ability.


\vspace{0.8mm} 
\noindent\textbf{Evaluation metrics.} Macro F1 is the most used metric given class imbalance in datasets, while accuracy is also popular. 
Some studies also consider precision, recall, and weighted F1 score.
\citet{eronen2022transfer} propose a new linguistic similarity metric based on the World Atlas of Language Structures (WALS) \cite{wals} to select optimal transfer languages for automatic HS detection.

\vspace{0.8mm} 
\noindent\textbf{Source code availability.} 
Out of 67 papers reviewed, 26 (\~38\%) made their source code available, whereas the remainder 41 (62\%) do not, hindering transparency and reproducibility. 

\begin{figure}[h]
  \centering
  \includegraphics[width=0.99\linewidth]{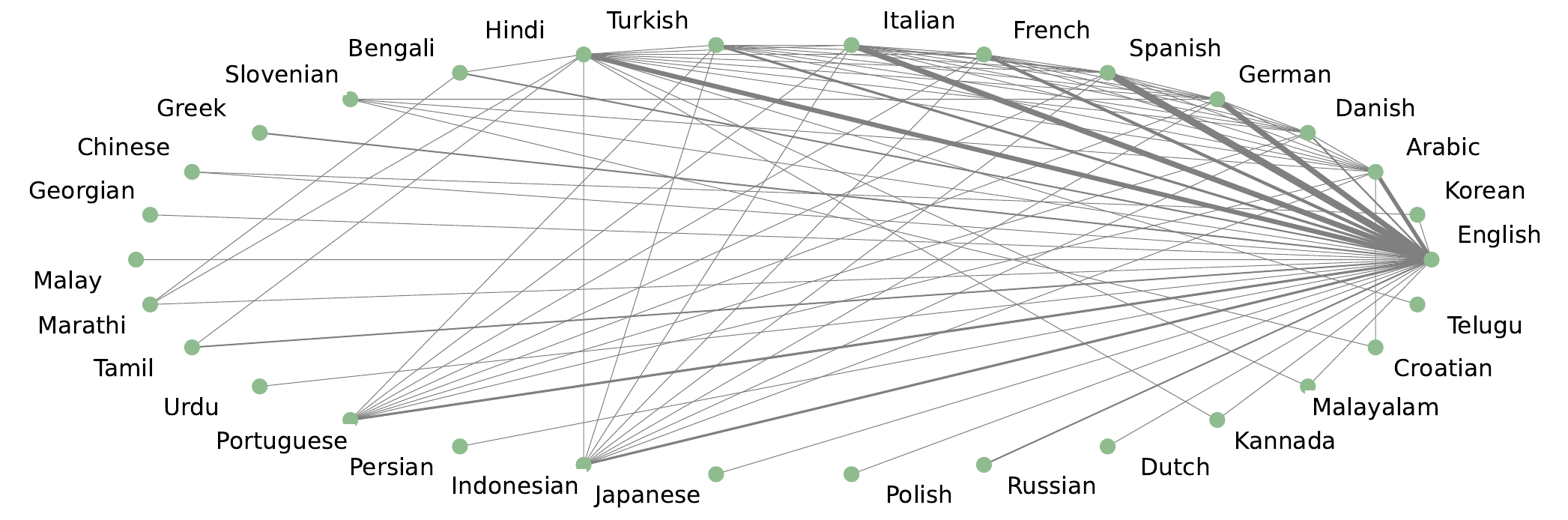}
  \caption{Synergies between languages. A link indicates that both languages have been used in a model, as source or target language, or to learn multilingual feature spaces. Thicker lines for higher frequencies.}
  \label{fig:lang-pair}
\end{figure}

\vspace{0.8mm} 
\noindent\textbf{Language pairs.} We present the frequency of language pairs in Figure \ref{fig:lang-pair}. English prevails as the source language, while the most frequent target languages are Spanish, German, and Italian. 
These languages share typological similarities with English, which likely facilitates their frequent pairing in cross-lingual tasks. 
This emphasises the impact of data availability and language similarity on the training and application of cross-lingual methods, and also reveals the limitations resulting from data scarcity.
Many studies have also explored cross-lingual learning by transferring English to Arabic, Hindi and French, as these languages have a relatively large amount of datasets available (see Figure \ref{fig:lang-family}).
Additionally, there are studies working on the transfer between non-English languages, among which Danish, Indonesian, Portuguese, and Turkish are frequent.

\vspace{0.8mm} 
\noindent\textbf{Comparative strengths and trade-offs.}  
While each transfer level offers viable solutions, their effectiveness can vary depending on the task setting and language pair characteristics.  
For example, \textbf{instance transfer} is often most effective when high-quality parallel corpora or reliable translation tools are available, enabling direct label projection or text translation. It tends to work well for typologically similar languages or when cultural and topical overlap is high.  
\textbf{Feature transfer} is particularly useful when parallel data is scarce but high-quality multilingual representations exist; it can capture shared semantic spaces while retaining language-specific nuances. This approach is generally more robust to moderate typological distance, but may struggle when the target language is under-represented in pre-training corpora.  
\textbf{Parameter transfer} offers the best flexibility across settings, but its success depends heavily on the pre-trained model coverage of the target language and the domain similarity between training and testing data. Parameter sharing in multilingual PLMs often generalises best within the same language family, while cross-family transfer may require additional adaptation (e.g., domain-specific fine-tuning or language-specific modules).  
In practice, selecting an appropriate transfer strategy requires balancing data availability, typological similarity between languages, and computational resources. Hybrid approaches that combine strategies can often yield the most reliable performance across diverse language families, such as using instance transfer for initial alignment and parameter transfer for fine-tuning. 

\section{Challenges and Future Directions}
\label{sec-challenge}

In this section, we examine key challenges encountered in cross-lingual offensive language detection in three aspects (language, dataset and approach), and then outline potential future directions.

\subsection{Language-Related Challenges}

\subsubsection{\textbf{Low-resource and Underrepresented Languages}}  
A central challenge is the limited availability of resources for many non-mainstream languages. As illustrated in Figure~\ref{fig:lang-pair}, the majority of studies focus on English as the source language and transfer primarily to typologically similar European languages (e.g., Spanish, German, French). 
In contrast, languages from other families (particularly African, Indigenous, and smaller Asian languages) remain underrepresented in both datasets and modelling efforts.  
This imbalance reflects broader trends in multilingual PLMs, where training corpora disproportionately favour high-resource languages~\cite{nozza2021exposing,rottger2022data}. As a result, CLTL systems often exhibit strong performance for well-represented languages but much weaker transferability for low-resource or distant languages~\cite{deshpande2022highly,eronen2022transfer}.  
The scarcity of labelled data, the uneven distribution of languages in PLMs, and the difficulty of transferring knowledge across typologically distant families represent major obstacles. 

\vspace{-1mm}
\subsubsection{\textbf{Diverse Linguistic Structures}}


Language diversity encompasses various aspects from basic grammatical rules to complex nuances unique to each language.
\emph{Linguistic distance} captures how typologically similar or different two languages are.
Transfer tends to be more successful across languages within a certain family that shares vocabulary, morphology, or syntactic patterns (e.g., English–German, Spanish–Italian), while performance often degrades for distant language pairs (e.g., English–Swahili, English–Chinese)~\cite{deshpande2022highly,eronen2022transfer}. 
Figure~\ref{fig:lang-pair} also demonstrates that cross-lingual experiments are concentrated within Indo-European languages, while distant language pairs are far less represented.  
This indicates that challenges for CLTL are not only the scarcity of data in low-resource languages, but also the uneven ability of models to generalise across varying degrees of linguistic distance. 
Furthermore, the presence of \emph{dialectal forms} within a single language amplifies its complexity. 
For example, the Arabic language is spoken by a wide range of countries across Asia, with complexities arising due to its multiple dialects \cite{husain2022tranfer}.

\vspace{-1mm}
\subsubsection{\textbf{Code Mixing}}

Non-English social media content is often characterised by code-mixing, where users blend languages, use transliterations, or incorporate multiple scripts within a single sentence or conversation, such as English-Hindi \cite{ibrohim2019translated,hande2021offensive} and English-Chinese \cite{jiang2023sexwes}. The English-Hindi (or Hinglish) combination appears in many cases, because the unique policies on regulating such speech in these regions add layers of complexity and make detection more challenging \cite{kapoor2019mind}.

\vspace{-1mm}
\subsubsection{\textbf{Cultural Variations}}

Culture is multifaceted and complex. 
The perception of offensive language can differ significantly across languages and cultures \cite{nozza2021exposing}.
What is considered non-offensive in one culture or language might be misinterpreted as signals of offense in another, leading to detection discrepancies.
Even within a single language, there can be cultural diversity.
For example, categorising English as a representative of ``western cultural background'' might ignore the differences between American and British cultures \cite{zhou2023cross}. 
Such cultural variations can hinder the effective transferability of language models \cite{rottger2022multilingual}.
Moreover, hateful expressions often employ figurative language, rhetorical figures and idioms, which are also language-dependent. This requires models to interpret underlying intent rather than relying solely on literal meanings \cite{pamungkas2021joint}.

\vspace{-1mm}
\subsubsection{\textbf{Explicit vs. Implicit Offensive Language}}  
Offensive language can be expressed explicitly or implicitly. Explicit abuse, such as direct insults or slurs, often transfers more easily across languages because it relies on surface lexical markers. In contrast, implicit abuse (expressed through sarcasm, stereotypes, or code-mixed language) is far more context-dependent and culturally situated. 
As a result, implicit forms are much harder to transfer from high-resource to low-resource settings, where contextual cues may differ significantly \cite{hartvigsen2022toxigen}.

\subsection{Dataset-Related Challenges}

\subsubsection{\textbf{Inconsistent Definition of Offensive Language}}

The diversity of HS encompassing from misogyny and racism to other forms of discrimination often leads to discrepancies in research \cite{nozza2021exposing}.
This has a bigger impact on low-resource languages \cite{bigoulaeva2021cross,bigoulaeva2022addressing,bigoulaeva2023label}, and introduces ambiguities that challenge the task, especially in multilingual scenarios \cite{vitiugin2021efficient}.
The lack of definition consistency across datasets in turn limits cross-lingual research, because many datasets in either source or target languages may be incompatible \cite{bigoulaeva2022addressing}.





\vspace{-1mm} 
\subsubsection{\textbf{Data Scarcity and Quality}}


\vspace{-1mm} 
\noindent\textbf{Limited Labelled Datasets.}
Low-resource languages often do not have the attention that major languages receive, leading to limited data availability \cite{burtenshaw2021dutch} and limited studies (see Figures \ref{fig:lang-family} and \ref{fig:lang-pair}).
Additionally, creating a high-quality labelled dataset requires extensive manual effort, which can potentially lead to high annotation and time costs as well as be harmful to annotators \cite{rottger2022data}.
Annotating offensive language is even more challenging due to the need for cultural and linguistic knowledge. 
Ethical and privacy concerns also arise when dealing with real-world data from social media platforms or other online sources \cite{jiang2022swsr}, because there are potential risks of exposing sensitive information or inadvertently promoting offensive language when studying it or trying to combat it.

\vspace{0.8mm} 
\noindent\textbf{Dataset Imbalance.}
Given that the majority of social media content is non-hateful, labels are typically skewed towards the non-hate label \cite{song2021study,bigoulaeva2022addressing,bigoulaeva2023label}. This skewness can lead to training issues, especially when working with small training corpora.
In addition, when considering multilingual or merged datasets, inter-language imbalance is another challenge \cite{deshpande2022highly}. 
Such datasets often display significant differences in the number of examples available for each language. This imbalance can bias models towards the semantic tendencies of languages that are overrepresented.

\vspace{0.8mm} 
\noindent\textbf{Dataset Bias.}
Biases such as topic, authorship, political affiliation, social media platform, and collection period can be one of the major issues and lead to potential generalisation challenges for cross-lingual models.
Among them, topic bias has been taken seriously in many works \cite{rottger2022data,arango2020hate,vitiugin2021efficient,rottger2022multilingual}.
While some datasets include general offensive language or HS, others might concentrate on specific issues such as immigrants, misogyny, politics, or religion.
Such topic-specific biases across datasets can harm model performance in cross-lingual classification \cite{pamungkas2021joint,stappen2020cross}.
Besides, the temporal aspect of data collection, affected by events in different periods, can further introduce bias, leading to datasets with varied topical foci \cite{stappen2020cross}.

\vspace{0.8mm} 
\noindent\textbf{Data Source Obsolescence.}
As online language rapidly transforms, especially influenced by moderation and real-world events, datasets can quickly become obsolete \cite{vitiugin2021efficient,alkhalifa2023building}. 
This obsolescence is further exacerbated by the emergence of new slang, metaphors, and colloquialisms that vary across languages and regions.
For instance, terms that normally have neutral meanings (such as donkey) may be used offensively in sexist text \cite{jiang2023sexwes}. 
The constrained and informal style of tweets, which are more like oral expressions than written language, complicates the preprocessing steps and leads to erroneous predictions \cite{pamungkas2020misogyny,aluru2021deep}. 
Emojis, associated with various forms of online harassment, further pose unique challenges \cite{corazza2020hybrid}.
Moreover, the domain and target of HS can shift significantly over time \cite{florio2020time}.
Real-world events, from local incidents to global crises like the COVID-19 pandemic, can give rise to new domains and terms of HS \cite{montariol2022multilingual}.
Then, it becomes increasingly challenging to create and continually update datasets customised for every possible language and domain, leading to frequent low-resource issues.

\vspace{-1mm}
\subsubsection{\textbf{Annotation Bias}}

Offensive language datasets can exhibit systematic gaps and biases in their annotations \cite{rottger2022multilingual}. 
First, different labelled datasets often utilise distinct annotation frameworks or strategies \cite{fortuna2018survey}. 
Second, the subjective nature of offensive language leads to tentatively disputed annotations among human annotators, where hateful content could be interpreted differently depending on annotators' understanding \cite{das2022data}. 
The annotation of these contents is easily influenced by individual sociodemographic and cultural backgrounds, often resulting in low inter-annotator agreement \cite{markov2021exploring}. 
Additionally, there is an ``annotation dilemma'' \cite{aluru2021deep}, whereby ambiguous instances might be annotated incorrectly by annotators but predicted accurately by the model.

\subsection{Approach-Related Challenges}

\subsubsection{\textbf{Limited Ability of Multilingual Pre-Trained Language Models}}

Cross-lingual models, especially multilingual PLMs, often lack generalisability and exhibit model bias \cite{pamungkas2021joint,hahn2021modeling,pelicon2021investigating}. 
The generalisability of transformer-based multilingual PLMs can be inconsistent, especially for typologically diverse languages, because they might be pre-trained with a highly focused set of languages or some languages with insufficient training samples \cite{deshpande2022highly}.
This can lead to bias and instability in model performance, depending on the model architectures, topical focus, linguistic distance, or cultural context in datasets \cite{mozafari2022cross,pamungkas2021joint}.
For example, PLMs may exhibit systematic bias when handling dialects, minority languages with various linguistic distances, and content towards marginalised communities. Since such varieties could be under-represented in pre-training corpora, PLMs tend towards standardised forms of dominant languages. This can result in degraded model performance, erasure of dialect-specific expressions, or misinterpretation of community-specific discourse. 
Model overfitting across target languages is another important issue.
Although multilingual models are less prone to overfitting on dataset-specific features than monolingual models, 
they will achieve poorer performance than monolingual ones in higher-resource settings, and they may require very different calibrations and adaptations across languages \cite{rottger2022data,rottger2022multilingual}.

\vspace{-1mm}
\subsubsection{\textbf{Cross-lingual Transfer Performance}}


CLTL has emerged as a promising solution and has shown effectiveness addressing data scarcity \cite{bigoulaeva2022addressing,sohn2019mc}. 
However, offensive language is deeply rooted in the specificity and diversity of language and culture, and its complexity poses significant obstacles to cross-lingual transfer, especially in zero-shot settings \cite{nozza2021exposing,montariol2022multilingual}.
Zero-shot transfer to multilingual training often suffers from performance deficiencies when compared to models trained on actual target language data \cite{pelicon2021investigating}.
It relies heavily on surface-level lexical and structural features, making them less effective when confronted with cultural contexts or implicit forms of abuse. This limitation has been consistently observed in empirical studies: zero-shot transfer yields significantly weaker results than target-language fine-tuning, even when only small amounts of labelled target data are available~\cite{pelicon2021investigating}. Moreover, cultural and linguistic variations across communities can further erode transferability, leading to biased or unreliable detection outcomes~\cite{nozza2021exposing}.

\vspace{-1mm}
\subsubsection{\textbf{Limitations of Machine Translation}}

Machine translation can help augment datasets and alleviate data scarcity issues. 
A prevalent practice is to utilise Google Translate API.\footnote{https://translate.google.com} \cite{aluru2021deep,pamungkas2021joint,jiang2021cross,elalami2022multilingual}
However, the effectiveness of cross-lingual detection models is directly linked to the quality and accuracy of the machine translation.
True semantics of the target text may change during the translation process.
Machine translation may diminish the toxicity degree and thereby reduce the perception of offensive content, especially in context-sensitive situations \cite{das2022data}.
Therefore, models that rely on such translations can experience performance degradation, especially when predicting instances with semantic shifts.
However, despite translation errors and uncertainties, they can still be valuable supplementary inputs for text classification \cite{sohn2019mc}.




\vspace{-1mm}
\subsubsection{\textbf{Poor Model Interpretability}}

While advanced deep learning-based models have shown good performance in detecting HS, they often operate as ``black boxes'' with a lack of transparency in the decision-making process \cite{aluru2021deep}.
This opacity makes it hard for humans to understand the underlying reasons for the model performance in a particular language or scenario, and analyse the model errors \cite{deshpande2022highly,vitiugin2021efficient}.
To bridge this interpretability gap, some tools are utilised to explain what models are doing, such as Local Interpretable Model-agnostic Explanations (LIME) \cite{lime} and SHapley Additive exPlanations (SHAP) \cite{shap}, and theoretical approaches from cognitive linguistics are also applied, e.g. frame semantics \cite{vitiugin2021efficient}.
A human-in-the-loop paradigm can also help by integrating human expertise into the model's learning process, but it can be difficult and uncertain for individuals to provide effective feedback to enhance the model performance \cite{vitiugin2021efficient}.

\subsection{Future Directions}
\label{sec:future}

This section outlines promising research directions and synthesises lessons from prior work. 

\vspace{-1mm}
\subsubsection{\textbf{Dataset Creation}}


A recurring finding is that \textbf{collecting even small amounts of target-language data} works well: fine-tuning with just a few hundred annotated examples consistently outperforms pure zero-shot transfer~\cite{rottger2022data}. Conversely, relying solely on zero-shot transfer does not work well, as it leads to significant drops in performance for typologically distant or culturally distinct languages~\cite{pelicon2021investigating}. Datasets should therefore be designed to cover \textbf{multiple languages, dialects, and cultural contexts}~\cite{ousidhoum2019multilingual}, while avoiding narrow topical bias from single platforms~\cite{rottger2022data}. Complementary benchmark resources such as XHATE-999~\cite{glavas2020xhate} and Multilingual HateCheck (MHC)~\cite{rottger2022multilingual} provide more reliable cross-lingual evaluation.

\vspace{-1mm}
\subsubsection{\textbf{Data Annotation}}
Research suggests that \textbf{incremental and selective annotation} works well: iteratively expanding small subsets increases efficiency while avoiding redundant labelling~\cite{rottger2022data}. 
In contrast, large-scale undirected annotation does not work well, as benefits quickly diminish relative to cost. 
To advance annotation quality, researchers should \textbf{recruit diverse annotators and provide culturally contextualised guidelines}, which mitigates annotation bias. 
Moreover, \textbf{active learning and semi-supervised strategies} have been shown to optimise annotator effort by focusing on ambiguous or uncertain cases~\cite{rottger2022data}. Expert feedback is particularly valuable for subtle and implicit abuse~\cite{vitiugin2021efficient}, but annotator well-being and ethical safeguards must also be prioritised, providing appropriate training and support to avoid potential harm.

\vspace{-1mm}
\subsubsection{\textbf{Integration of Additional Features}}

Studies consistently show that integration of additional features can also be effective.
Adding \textbf{language-agnostic signals such as typographic features} (capitalisation, punctuation, emojis) works well~\cite{kumar2022multi,markov2021exploring,corazza2020hybrid}. Emojis, in particular, often carry stable meanings across languages and provide additional emotional cues that complement textual signals. 
\textbf{Stylometric features} such as function word usage, sentence length, or character-level distributions can also strengthen detection, since they capture writing style cues of toxic content that can be correlated with specific social media users ~\cite{markov2021exploring}.   
Similarly, \textbf{domain-specific resources} such as HurtLex and tailored embeddings work well in capturing hate-related semantics across languages~\cite{pamungkas2020misogyny,jiang2023sexwes}. For example, multilingual lexicons of offensive terms have been shown to boost recall by reducing false negatives, while domain-aware embeddings such as sexism-specific representations (SexWEs) provide finer-grained distinctions for subtle or implicit abuse.  
Relying only on general-purpose multilingual embeddings does not perform well, which often misses culture-specific or implicit abuse patterns~\cite{arango2021cross}. They may capture general semantic similarity but struggle with figurative expressions, reclaimed slurs, or locally grounded derogatory terms. Over-reliance on generic embeddings risks reinforcing biases by ignoring context-specific linguistic cues.  
Furthermore, infusing \textbf{cultural features} underscores the significance of language-specific knowledge for geographically sensitive tasks.
The creation of culturally-aware models, informed by multidisciplinary studies involving anthropologists and sociologists, can provide a richer understanding of cultural dimensions in hateful content \cite{zhou2023cross}. 
Such work can produce different forms of cultural features, such as switching pattern matrix \cite{bansal2020code}, bilingual language pairs \cite{jiang2023sexwes}, cross-cultural similarities \cite{arango2021cross}, and social dynamics among users \cite{nozza2021exposing}.

\vspace{-1mm}
\subsubsection{\textbf{Multilingual Pre-Trained Language Models}}

Multilingual PLMs can be applied to low-resource target languages without further training, thus bridging the data availability gap between high- and low-resource languages.
However, their large number of parameters carries significant computational costs during training and fine-tuning \cite{pikuliak2021cross}.
Future research is geared towards optimising these multilingual PLMs for enhanced efficiency and interpretability.
Additionally, two strategies are emerging to improve the generalisability and scalability of multilingual PLMs so that they can scale to handle large datasets or multiple languages simultaneously \cite{arango2020hate}.
The first emphasises pre-training models using \textbf{data from relevant sources or language communities} (such as social platforms and hateful domains), as demonstrated by XLM-T \cite{montariol2022multilingual} and AbuseXLMR \cite{gupta2022multilingual}. This can mitigate model bias across language communities. 
The second strategy focuses on models based on \textbf{specific low-resource languages, dialects or those from similar language families}, as seen with MuRIL for Indic languages \cite{das2022data} and AraBERT for dialects \cite{husain2022tranfer}, which requires inclusion of dialectal and under-represented language data in both training and evaluation.
Meanwhile, it is important to \textbf{evaluate transfer effectiveness across both similar and distant language pairs}, rather than relying solely on high-resource or typologically similar examples. Otherwise, multilingual PLMs risk reinforcing existing performance gaps, performing well for Indo-European languages while neglecting typologically distant or underrepresented languages.

\vspace{-1mm}
\subsubsection{\textbf{Cross-lingual Training Strategies}}

Various cross-lingual training strategies have been proven effective.
\textbf{Adaptive training techniques} can dynamically adjust to the specificities of languages and dialects, as well as bridge the gap between offensive language detection and PLMs. 
By leveraging meta-learning or few-shot learning, models can be trained to quickly adapt to new low-resource languages with limited labelled data \cite{mozafari2022cross}. 
Furthermore, collaborative \textbf{multi-task learning} can make models more generalisable in cultural variations and resistant to overfitting by training them and sharing representations on multiple auxiliary tasks simultaneously across languages or similar NLP tasks \cite{montariol2022multilingual}.
The use of \textbf{adversarial training}, where models are trained against adversarial examples, can also enhance their robustness and generalisation. This is especially important for offensive language, as attackers often use subtle language tricks to obfuscate and bypass detection \cite{shi2022cross,jin2020bert,ryzhova2022training}.
Moreover, \textbf{cascade learning} can greatly bolster model performance and enhance out-of-domain generalisability by initially training models on source data or more diverse datasets and subsequently fine-tuning the results by simultaneously or progressively incorporating target language data  \cite{rottger2022data,pelicon2021investigating}.
Besides, \citet{vitiugin2021efficient} highlight that \textbf{human feedback} can be of great value to help detect subtleties of abuse and phrases during model training, \citet{ranasinghe2021multilingual} explore \textbf{language-specific preprocessing} like segmentation in morphologically rich languages (e.g., Arabic, Turkish), and \textbf{multi-source training}, with augmented datasets, is also able to capture the diversities and commonalities of different linguistic patterns \cite{hande2021benchmarking,markov2021exploring,chiril2019multilingual}.
Future research should focus on systematically comparing these strategies and identifying which are most effective under varying resource and language-family conditions.

\vspace{-1mm}
\subsubsection{\textbf{Application of Large Language Models}}


With the emergence of Large Language Models (LLMs), the landscape of offensive language detection has rapidly shifted. Models such as GPT-3 \cite{brown2020language}, GPT-4 \cite{openai2023gpt4}, Fine-tuned LAnguage Net (FLAN) \cite{wei2021finetuned} have demonstrated impressive multilingual capabilities, and recent surveys have highlighted their growing importance~\cite{zhao2023survey}.
\textbf{Instruction-tuned LLMs with prompting} can already perform competitive detection across monolingual and multilingual settings without the need for task-specific fine-tuning~\cite{plaza2023respectful}. Prompt design strategies, such as incorporating \textbf{task descriptions or input-label demonstrations}, further enhance performance~\cite{chen2022adaprompt,han2022designing,roy2023probing}.
LLMs also show promise in addressing data scarcity by \textbf{generating synthetic examples} for low-resource languages~\cite{hartvigsen2022toxigen}, and in \textbf{providing explanations} that improve interpretability~\cite{wang2023evaluating}.  
Despite their advantages, LLMs are not a full replacement for prior methods. They often underperform specialised multilingual PLMs on fine-grained tasks when little prompting expertise is available, and they remain costly to deploy at scale. Traditional approaches, such as leveraging domain-specific lexicons (e.g., HurtLex) or designing language-family-specific PLMs (e.g., MuRIL, AraBERT), still \emph{work well} for targeted low-resource or domain-specific scenarios, offering more efficiency and control.  

Overall, while early work on LLMs in this area was scarce at the time of our survey design, the rapid pace of development since then underscores their growing importance.
Future studies should explicitly and systematically compare LLM-based approaches with prior CLTL techniques to establish where LLMs offer clear advantages and where CLTL methods remain more effective.
Another important research direction is to explore the complementarity between LLMs and CLTL approaches. For example, lexicon features, linguistic distance, and culturally-aware representations can be integrated as prompts or external knowledge sources to guide LLMs, while smaller multilingual PLMs can be lightweight alternatives in resource-constrained environments.

\section{Conclusion}
\label{sec-conclusion}

This survey provides a detailed exploration of CLTL techniques in detecting offensive language online,
Our survey is the first to specifically focus on cross-lingual detection, analysing a collection of 67 papers. 
We summarise the characteristics of multilingual offensive language datasets and cross-lingual resources.
We further emphasise the diversity and complexity of CLTL techniques, highlighting the varying levels of transfer --instance, feature, and parameter-- and popular strategies to improve performance. 
The survey also reveals several ongoing challenges in cross-lingual offensive language detection, such as linguistic diversity, dataset scarcity, and model generalisability. 
We also point out potential research opportunities, emphasising the need for new datasets across low-resource languages, as well as more robust, generalisable, and ethically sound CLTL approaches. 
Our survey highlights the importance of advancing CLTL approaches, which can serve as both a key reference for current practices and a guide for future research in this evolving field.


\section*{Limitations}
While this survey provides a comprehensive synthesis of CLTL-based offensive language detection, several limitations remain.

\begin{itemize}
    \item \textbf{Focused scope:} We focus exclusively on cross-lingual text classification for offensive language detection, excluding related modalities (e.g., multimodal detection) and other transfer scenarios such as cross-domain or cross-platform learning.
    \item \textbf{Coverage of literature:} Our selection presents a comprehensive review of the literature which is updated up to the cut-off point of July 2023.
    \item \textbf{Benchmark comparison:} Although we summarise performance trends, we do not conduct benchmarking experiments with reported results, which may vary due to evaluation setups.
\end{itemize}

\section*{Ethical Considerations}
\label{sec-ethic}
CLTL-based offensive language detection systems can raise several ethical considerations with the datasets, methods, and challenges we have discussed.

\begin{itemize}
    \item \textbf{Bias in multilingual datasets:}  
    Multilingual offensive language datasets often suffer from topic bias, imbalanced language representation, and inconsistent annotation guidelines, which can propagate through CLTL models across languages and communities. Balanced and culturally diverse data collection, transparent reporting of dataset composition, and fairness-aware evaluation are needed.

    \item \textbf{Cultural sensitivity:}  
    Offensive expressions are often context-dependent, and models trained on data from one cultural context may misclassify content in another, e.g., sarcastic expressions such as ``well done'' or idioms like calling someone a ``clever fox'' may be benign in one language but misinterpreted as neutral or offensive in another cultural context when translated.
    Culturally-aware annotation and evaluation with native speakers from target communities are necessary.

    \item \textbf{Model transparency and interpretability:}  
    CLTL systems, particularly those based on PLMs, often act as``black boxes''. This lack of interpretability can hinder error analysis. Techniques such as attention or post-hoc visualisations (e.g., LIME, SHAP), and clear model training and evaluation setups can improve transparency.

    \item \textbf{Responsible use and risk mitigation:}  
    While CLTL-based systems can support harm reduction, they also carry risks of misuse for censorship, political suppression, or disproportionate moderation of certain language groups. Deployment should have clear governance frameworks, stakeholder engagement, and human-in-the-loop review mechanisms.
\end{itemize}

\begin{acks}
Aiqi Jiang is funded by China Scholarship Council (CSC Funding, No. 201908510140). Arkaitz Zubiaga acknowledges support from the European Union and UK Research and Innovation under Grant No. 101073351 as part of Marie Skłodowska-Curie Actions (MSCA Hybrid Intelligence to monitor, promote, and analyse transformations in good democracy practices).
\end{acks}

\bibliographystyle{ACM-Reference-Format}
\bibliography{reference}


\end{document}